\documentclass[lettersize,journal]{IEEEtran}
\usepackage{easyReview}

\usepackage{amsmath,amsfonts}
\usepackage{subfigure}
\usepackage{algorithmic}
\usepackage{algorithm}
\usepackage{array}
\usepackage[caption=false,font=normalsize,labelfont=sf,textfont=sf]{subfig}
\usepackage{textcomp}
\usepackage{stfloats}
\usepackage{url}
\usepackage{verbatim}
\usepackage{graphicx}
\usepackage{cite}
\usepackage{arydshln}

\usepackage{rotating}
\usepackage{makecell}
\newsavebox{\tablebox}
\usepackage{siunitx}

\usepackage{amsthm}

\newtheorem{remark}{Remark}

\newtheorem{theorem}{Theorem}
\newtheorem{myDef}{Definition}

\newcommand{\iq}{\mathbf{i}}
\newcommand{\jq}{\mathbf{j}}
\newcommand{\kq}{\mathbf{k}}
\newcommand{\Aq}{\mathbf{A}}
\newcommand{\aq}{\mathbf{a}}
\newcommand{\Dq}{\mathbf{D}}

\newcommand{\xq}{\mathbf{x}}

\newcommand{\zq}{\mathbf{z}}
\newcommand{\Kq}{\hat{K}}

\newcommand{\Uq}{\mathbf{U}}
\newcommand{\uq}{\mathbf{u}}

\newcommand{\vq}{\mathbf{v}}

\newcommand{\wq}{\mathbf{w}}

\newcommand{\sq}{\mathbf{s}}

\newcommand{\qq}{\mathbf{q}}
\newcommand{\pq}{\mathbf{p}}

\newcommand{\diag}{\operatorname{diag}}

\usepackage{epsfig,amsmath,bm,epstopdf}

\def\SVTV{{\rm{SVTV}}}

\def\CTV{{\rm{CTV}}}
\def\MTV{{\rm{MTV}}}

\def\CSTV{{\rm{CSTV}}}

\def\GMRES{{\rm{GMRES}}}
\def\QGMRES{{\rm{QGMRES}}}

\def\Aq{{\bf A}}

\def\aq{{\bf a}}

\def\bq{{\bf b}}

\def\Iq{{I}}

\def\Pq{{\bf P}}
\def\pq{{\bf p}}
\def\Qq{{\bf Q}}
\def\qq{{\bf q}}

\def\rqq{{\bf r}}
\def\sq{{\bf s}}

\def\vq{{\bf v}}

\def\xq{{\bf x}}

\begin{document}
	
	\title{A New Cross-Space Total Variation Regularization Model for Color Image Restoration with\\ Quaternion Blur Operator\thanks{This work is  supported  in part by the  National Key Research and Development Program of China under grant 2023YFA1010101; the National Natural Science Foundation of China under grants 12171210, 12090011,  61971234, 12271467 and 11771188;  the “QingLan” Project for Colleges and Universities of Jiangsu Province (Young and middle-aged academic leaders);
the Major Projects of Universities in Jiangsu Province under grant 21KJA110001;    the Natural Science Foundation of Fujian Province of China under grant 2022J01378;  and the Scientific Research Foundation of NUPT under grant NY223008.
M. Ng’s research is partly supported by the National Key Research and Development Program of China under Grant 2024YFE0202900; HKRGC GRF 17201020 and 17300021, HKRGC CRF C7004-21GF, and Joint NSFC and RGC N-HKU769/21.
(Corresponding author: Michael K. Ng.)}
}
	
	\author{Zhigang Jia\thanks{Zhigang Jia is  with the School of Mathematics and Statistics \& RIMS, Jiangsu Normal University, Xuzhou 221116, P.R. China (e-mail: zhgjia@jsnu.edu.cn).},
\and Yuelian Xiang\thanks{Yuelian Xiang is  with the School of Mathematics and Statistics, Jiangsu Normal University, Xuzhou 221116, P.R. China and  also  with the General Education School,  Wuhan Vocational College of Software and Engineering, Wuhan 430205, P.R. China (e-mail:  yuelianx@126.com).},
\and 	 Meixiang Zhao\thanks{Meixiang Zhao is  with the School of Mathematics and Statistics, Jiangsu Normal University, Xuzhou 221116, P.R. China (e-mail: zhaomeixiang2008@126.com).},
\and Tingting Wu\thanks{Tingting Wu is with the School of Science, Nanjing University of Posts and Telecommunications, Nanjing 210003, P.R. China (e-mail: wutt@njupt.edu.cn).},
\and Michael K. Ng\thanks{Michael K. Ng is with the Department of Mathematics, Hong Kong Baptist University, Hong Kong (e-mail: michael-ng@hkbu.edu.hk).}
	}
	
	\markboth{Journal of \LaTeX\ Class Files,~Vol.~14, No.~8, February~2024}%
	{Shell \MakeLowercase{\textit{et al.}}: 
A New Cross-Space Total Variation Model 
}
	
	
	\maketitle
	
	\begin{abstract}The cross-channel deblurring  problem in color image processing is difficult to solve due to the complex coupling and structural blurring of color pixels. Until now, there are few efficient algorithms that can reduce color artifacts in deblurring process. To solve this challenging problem,  we present a novel cross-space total variation (CSTV) regularization model for color image deblurring by introducing  a quaternion blur operator and a  cross-color space regularization functional.  The existence and uniqueness of the solution are proved and a new L-curve method is proposed to find a  balance of regularization terms on different  color spaces.  The Euler-Lagrange equation is derived to show that CSTV has taken into account the coupling of all color channels and the local smoothing within each color channel.  A quaternion operator splitting method is firstly proposed to enhance the ability of color artifacts reduction of the CSTV regularization model. This strategy also applies to the well-known color deblurring models. Numerical experiments on  color image databases  illustrate the efficiency and effectiveness of the new model and algorithms. The  color images restored by them successfully maintain the color and spatial information and are of higher quality  in terms of  PSNR, SSIM, MSE and CIEde2000 than the restorations of the-state-of-the-art methods. 
	\end{abstract}
	

\section{Introduction}
\IEEEPARstart{C}ross-channel deblurring  is  one of the most  important topics in color image processing.  It is difficult to solve due to the complex coupling and structural blurring of color pixels that are difficult to characterize.   
The classic and advanced color image total variation (TV) regularization  restoration models,  consisting of  regularization and  fidelity terms, have made outstanding contributions to denoising and deblurring. However,  their innovation is focused on the improvement of TV regularization term on one color space and there is still a lack of efficient algorithms that can reduce color artifacts.  A long-overlooked, but very  important, requirement to develop advanced models is to define color TV regularization functional on different color spaces and develop algorithm that preserves the coupling of color pixels.  
 In this paper, we  present a novel cross-space TV (CSTV) regularization model for color image restoration that achieves such requirement.  Moreover, we develop a general quaternion operator splitting algorithm to enhance the  ability of reducing color artifacts of color image deblurring models.

A groundbreaking color image regularization  is  the  (global)  channel coupling color TV \cite{bc98},  followed by a local version  \cite{brc08}. These two regularization functionals couple red (R), green (G) and blue (B) channels by different norms of TVs on three channels, that is,  norms of vectorial TV of color image in the RGB color space.  A general framework, called collaborative TV   \cite{dms16},  incorporates various norms along different dimensions to provide a comprehensive approach to regularization and the considered norms includes nuclear, Frobenius and spectral norms.    Recently,  Duan et al. \cite{dztg22}  proposed a new Beltrami regularization model for color image denoising and an efficient and robust operator splitting method, with regarding color images as manifolds embedded in a five dimensional spatial-chromatic space.   And a  new color elastica model is developed in \cite{ltkg21,ltkg23} by using the Polyakov action and a Laplace--Beltrami operator on color channels.   These beautiful regularization functionals  greatly enriched color image models based on vector representation in one color space. 

To find more efficient manners of coupling, various studies explore color space transformation methods to generate  regularization functionals on new color space. The chrominance, luminance, or R/B component was formed through a weighted linear combination of the R, G and B channels of the color image in \cite{cks01}. This process is also represented as $\sum_{k}~TV(u_k \circ \psi)$ in \cite{dms16}, where $\psi$ is an orthonormal transform to describe the color space transformation.  Besides, the human eye is highly sensitive to changes in the opponent color channel. In \cite{sy14},  the opponent transform is employed to transform the RGB color space into the opponent space and the decomposition of the coupling of color channels is allowed.   Recently, a novel saturation-value TV (SVTV) regularization model is introduced in \cite{jnw19} based on quaternion representation of color images and the existence and uniqueness of the solution are proved. Such SVTV regularization functional successfully reflects the physical principle of the human visual system and considers the coupling of color channels. So SVTV reconstructs color images of high quality and with  slight color artifacts.  In these years,  the SVTV regularization functional has been combined with other methods to handle other image processing tasks and the new models  achieve at a high level on numerical performance; see \cite{hmwz21,wpm20,ws22,wym22} for instance. Above models consider color image restoration in a single color space and their regularization functionals have been well known for their  prior on color edges.

Unlike the successful improvement of regularization terms,   the development of fidelity terms of color image restoration models has been hindered for a long time by  the difficulty of characterizing complex coupling and structured blur of color pixels and the hardness to eliminate  color artifacts. 
A new understanding of the blurring process is in need.  
Recall that a general color image degradation model   (see \cite{fnb06a,fnb06b,gkch91} for instance) is 
\begin{equation}\label{e:imgdeg}
	z(x,y)=\hat{K} \star u(x,y) + n(x,y),
\end{equation}
where $z(x,y),\hat{K},u(x,y),n(x,y)$ denote degraded color image, blur operator,  original color image and additive noise, respectively, and $(x,y)$ refers to their pixel coordinates.
Conventionally,  the blur operator  $\hat{K}$ is a three-by-three block matrix consisting of several two-dimensional convolution operators \cite{gkch91} and the fidelity functionals are defined by different norms of $\hat{K} \star u(x,y)-z(x,y)$.   
In order to make progress, we use  quaternions to represent  color pixels and introduce a novel quaternion convolution operator $\Qq$ (defined in Section \ref{ss:fidelity}) to characterize the cross-channel blurring process. This blur operator $\Qq$  depicts the mutual artifacts between color channels.   It can be seen as a splitting factor of $\hat{K}$ and is used to improve the fidelity term.

To the best of our knowledge, there are still no cross-color space mathematical model with strict proof of existence and uniqueness of the solution for color image cross-channel deburring problem.  In this paper, we present a novel color image restoration model to solve the cross-channel deblurring problem and theoretically analyze its solvability.  The contribution is in three aspects:
\begin{itemize}

	\item  A CSTV regularization model is presented for color image restoration by introducing a novel quaternion blur operator and a new regularization functional.
The CSTV regularization functional is defined on different color spaces and preserves the coupling of color pixels in the deblurring process.  This  generalizes the well-known TV regularization terms from one color space to  two or more color spaces.

\item A new quaternion operator splitting algorithm is developed to enhance the  ability of reducing color artifacts in deblurring process of the proposed CSTV regularization model. This quaternion operator splitting algorithm is universal and  can be applied to improve classical color image deblurring models.

\item  The newly proposed models and algorithms are applied to solve the cross-channel deblurring problems of natural color images.  Their efficiency and superiority are indicated by numerical results  in terms of visual, PSNR, SSIM, MSE and CIEde2000 criteria.   These numerical results support the assertion that the proposed new methods can better preserve the color fidelity and texture.
\end{itemize}

This paper is organized as follows. 
In Section \ref{s:prilim}, we recall  preliminary information about quaternions and  color TV regularization functionals.
In Section \ref{s:model}, we propose a novel CSTV regularization  model for color image restoration and  the corresponding existence and uniqueness theory of the solution. 
In Section \ref{s:numalg}, we present an  effective  quaternion operator splitting algorithm to solve the proposed CSTV regularization model. 
In Section \ref{s:experiments}, we demonstrate numerical examples to illustrate the superiority of the proposed methods. 
In Section \ref{s:conclusion}, we present  concluding remarks.

\section{Preliminaries}\label{s:prilim}
In this section, we shortly review quaternions and the existing TV regularization functionals  for color image processing.

Let $\mathbb{Q}=\left\{a_0+a_1\iq+a_2\jq+a_3\kq~|~a_0,a_1,a_2,a_3\in \mathbb{R}\right\}$ denote the quaternion skew-field, $\mathbb{Q}^{n}$  the  set of $n$-dimensional quaternion vectors, and $\mathbb{Q}^{m\times n}$  the  set of $m\times n$ quaternion matrices \cite{Hamilton66}, where three imaginary units $\iq, \jq, \kq$ satisfy
$$\iq^2=\jq^2=\kq^2=\iq\jq\kq=-1.$$
 For a quaternion $\aq=a_0+a_1\iq+a_2\jq+a_3\kq$, $a_0$ is called real part  and $a_1,a_2,a_3$ are called three imaginary parts. Pure quaternion is  nonzero quaternion with zero real part. The conjugate and modulus of $\aq$ are defined by
$\bar{\aq}=a_0-a_1\iq-a_2\jq-a_3\kq$ and $|\aq|=\sqrt{\bar{\aq}\aq}=\sqrt{a_0^2+a_1^2+a_2^2+a_3^2}$, respectively.
Every nonzero quaternion is invertible and its unique inverse is defined by $\aq^{-1}=\bar{\aq}/|\aq|^2$.
The conjugate transpose of  quaternion vector 
$\vq=v_0+v_1\iq+v_2\jq+v_3\kq\in\mathbb{Q}^{n}$ 
 is defined as
$\vq^*=v_0^T-v_1^T\iq-v_2^T\jq-v_3^T\kq$, where $v_0,v_1,v_2,v_3\in\mathbb{R}^{n}$  and  $\cdot^T$ denotes the transpose operator.
Similarly, the conjugate transpose of quaternion matrix $\Aq=A_0+A_1\iq+A_2\jq+A_3\kq\in\mathbb{Q}^{m\times n}$  is defined as
$\Aq^*=A_0^T-A_1^T\iq-A_2^T\jq-A_3^T\kq$, where $A_0,A_1,A_2,A_3\in\mathbb{R}^{m\times n}$.
The inner product of two quaternion vectors, $ \uq=[\uq_{i}],~\vq=[\vq_{i}] \in \mathbb{Q}^{n}$, is defined as $ \left<\uq,\vq\right>=\sum_{i=1}^{n}\vq_{i}^{*}\uq_{i} $.
According to \cite{jing21}, we define a homeomorphic mapping $\Re$ from quaternion   matrices, vectors, scalars or operators to structured real matrices or operators: 
\begin{equation}\label{d:hmapping}
\Re(\Aq)=\left[\begin{array}{rrrr}
			A_0&-A_1&-A_2&-A_3\\
			A_1&A_0&-A_3&A_2\\
			A_2&A_3&A_0&-A_1\\
			A_3&-A_2&A_1&A_0
		\end{array}\right]\!\!.
\end{equation}
The inverse mapping of $ \Re $ on the structured real matrices is defined by $ \Re^{-1}(\Re(\Aq))=\Aq $.
Let $\Re_c(\Aq)$ denote the first column of $\Re(\Aq)$.  
The inverse mapping of $ \Re_c $ is similarly defined by $ \Re_c^{-1}(\Re_c(\Aq))=\Aq $.

Now we introduce the  measurement of quaternion vectors and matrices. The absolute of quaternion vector $\vq=[\vq_i]\in\mathbb{Q}^n$  is 
$|\vq|=[|\vq_i|]\in\mathbb{R}^n.$ Similarly, the absolute of quaternion matrix $\Aq=[\aq_{ij}]\in\mathbb{Q}^{m\times n}$  is  
$|\Aq|=[|\aq_{ij}|]\in\mathbb{R}^{m\times n}.$  From \cite{j19}, we know that quaternion vector (or matrix) norms are   functions from quaternion vectors (or matrices) to nonnegative real numbers.  
\begin{myDef}\label{d:pnorms} Let $p\ge 1$. The $p$-norm of $\vq\in\mathbb{Q}^{n}$  is
	$\| \vq \|_p=\big(\sum_{i=1}^n |\vq_{i}|^p\big)^{\frac{1}{p}}.$
	The $p$-norm and $F$-norm of  $\Aq \in\mathbb{Q}^{m\times n}$ are 
	$$\|\Aq\|_p=\max_{\xq\in\mathbb{Q}^{n}/\{0\}}\frac{\|\Aq\xq\|_p}{\|\xq\|_p},~\|\Aq\|_F=\big(\sum\limits_{i=1}^m\sum\limits_{j=1}^n |\aq_{ij}|^2\big)^{\frac{1}{2}}.$$
\end{myDef}
\noindent
From Definition \ref{d:pnorms}, one can easily derive that $\|\vq\|_2=\|\Re_c(\vq)\|_2=\frac{1}{2}\|\Re(\vq)\|_2$,   $\|\Aq\|_2=\|\Re(\Aq)\|_2$ and $\|\Aq\|_F=\|\Re_c(\Aq)\|_F=\frac{1}{2}\|\Re(\Aq)\|_F$.

Next, we give a brief description of the SVTV regularization functional \cite{jnw19}.  A  pure imaginary quaternion function
\begin{equation}\label{e:uhatu0}\uq(x,y)=u_1(x,y)\iq+u_2(x,y)\jq+u_3(x,y)\kq \end{equation}
 was used to represent a color image  in the RGB color space  in  \cite{jnw19}, 
where $(x,y)$ denotes the position of a color pixel in a given range $\Omega$ and three real bivariate functions $u_{i}(x,y)$ $(i=1,~2,~3)$ denote pixel values of R, G and B channels,  respectively.  
 For convenience,  we define a new transformation map  $\mathcal{T}$ as  
 \begin{equation}\label{Tfun}
 \mathcal{T}(\qq(x,y)):=q(x,y)=[q_1(x,y), q_2(x,y),q_3(x,y)]^T
 \end{equation}
 for $\qq(x,y)=q_0(x,y)+q_1(x,y)\iq+q_2(x,y)\jq+q_3(x,y)\kq$.
A real-valued version of SVTV was also given in \cite{jnw19}, which makes it possible to  handle by real calculations instead of quaternion calculations.
 For simplicity, let $\partial_{x/y}$  denote $\partial_{x}$ or $\partial_{y}$. 
 Define  $\partial_{x/y}\uq(x,y)=\partial_{x/y}u_1(x,y)\iq+\partial_{x/y}u_2(x,y)\jq+\partial_{x/y}u_3(x,y)\kq$ and
\begin{equation} \label{matrix:C}
	C=\left[
	\begin{array}{rrr}
		2&-1&-1\\
		-1&2&-1\\
		-1&-1&2
	\end{array}
	\right]\!\!.
\end{equation}
Let $u(x,y)=\mathcal{T}(\uq(x,y)).$
The saturation and value components of color image $\uq(x,y)$ are defined as
\begin{equation}\label{e:sv_norm}
|\uq(x,y)|_s\!\!=\!\!\frac{1}3\|Cu(x,y)\|_2,
|\uq(x,y)|_v\!\!=\!\!\frac{1}{\sqrt{3}}|\sum\limits_{i=1}^3u_i(x,y)|.
\end{equation}
The SVTV regularization  functional is defined in \cite{jnw19} by
\begin{equation}\label{e:svtv}
	\begin{aligned}
		\SVTV(\uq(x,y))=\int_\Omega\sqrt{|\partial_x \uq(x,y)|_s^2+|\partial_y \uq(x,y)|_s^2}\\
		+\alpha\sqrt{|\partial_x \uq(x,y)|_v^2+|\partial_y \uq(x,y)|_v^2}dxdy,
	\end{aligned}
\end{equation}
where $\alpha$ is a weighting parameter.  This is  exactly a  functional of real functions  $u_1(x,y),~u_2(x,y)$ and $u_3(x,y)$.  

Before  SVTV regularization functional, there are several well-known TV regularization functionals developed for color image in the RGB color space. For instance,  
\begin{itemize}
\item Blomgren and Chan \cite{bc98}:
\end{itemize}
\begin{equation} \label{ctv1}
{\rm CTV}_{1}(u(x,y)) := \sqrt{\sum_{i=1}^{3} \bigg(\int_{\Omega}\|\nabla u_i(x,y)\|_2~dxdy\bigg)^2}.
\end{equation}
\begin{itemize}
\item Bresson and Chan \cite{brc08}:
\end{itemize}
\begin{equation} \label{ctv2}
{\rm CTV}_{2}(u(x,y)) := \int_{\Omega}\sqrt{\sum_{k=1}^{3}\|\nabla u_i(x,y)\|_2^2}~dxdy.
\end{equation}
\begin{itemize}
\item Sapiro  \cite{Sapiro1996}:
\end{itemize}
\begin{equation}\label{e:VTV}
{\rm VTV}(u(x,y))\! :=\! \int_{\Omega} \! f( \lambda ( (\nabla u(x,y))^T\nabla u(x,y) )) dxdy,
\end{equation}
where
$\lambda(\cdot)$ refers to  eigenvalues and $f(\cdot)$ is a penalty function on eigenvalues.
In the above equations,   $\nabla=[\partial_x,\partial_y]^T$ represents the gradient operator.
In  \cite{dms16}, classic and advanced versions of  vector TV regularization functionals are summarized into a classification called  collaborative total variation (CTV) regularization functional. When applied to color images, they are all based on one color space.  
A new regularization on different color spaces will be firstly introduced in this paper.

\section{Cross-space total variation regularization model for color image restoration}\label{s:model}

In this section, we  present a novel color image restoration model to solve the cross-channel deblurring problem.

With the degradation model \eqref{e:imgdeg} in hand,  we  are concerned with a  general framework for color image restoration model
	\begin{equation}\label{e:gene}
		\mathop{\min}\limits_{
			\bm{\lambda} \in \mathbb{R}^m
			\atop
			u(x,y) \in \mathbb{BV}(\Omega)
		}~J(u,\lambda)+F(u,z,\hat{K}),
	\end{equation}
where $\mathbb{R}^m$ denotes the $m$-dimensional real-valued vector space, $\mathbb{BV}(\Omega)=\{u(x,y)= 
[ u_1(x,y), u_2(x,y), u_3(x,y) ]^T~|~$ $u_i(x,y)\in {\rm BV}(\Omega),~(x,y)\in\Omega\}$  and ${\rm BV}(\Omega)$ is bounded variation space in a given region $\Omega\subseteq\mathbb{R}^2$.
The functional $F(u,z,\hat{K})$ is a data fidelity term to control the distance between the targeted and observed  color images under the blurring operator.
 The functional $J(u,\lambda)$ punishes the gradients of the targeted color image $u(x,y)$  in different color spaces and it is called a cross-color space regularization term with regularization parameters ${\lambda}=(\lambda_1,\lambda_2,\cdots,\lambda_m)\in\mathbb{R}^m$.

\subsection{Fidelity functional with quaternion blur operator}\label{ss:fidelity}

The fidelity functional $F(u,z,{\hat K})$  aims to minimize the distance between the observed color image $z$ and the original color image $u$ after a blurring operation. 
In this subsection,  we introduce a new characterization of  blur  process  by quaternion operator, which leads to an 
algorithm to compute a solution without color artifacts. 
At first, we recall the cross-channel blur mechanism. 
The color image cross-channel blurring process is  mathematically described as
\begin{align}\label{e:blurmodel0}
	\hat{u}(x,y)& = \hat{K} \star u(x,y),
\end{align}
where
\begin{equation}\label{e:KW}
\begin{aligned}
&\qquad \qquad\qquad\qquad \hat{K} =W\odot K=[K_{ij}\omega_{ij}]_{3\times 3},\\
	&K\!\!=\!\!\left[\!
	\begin{array}{ccc}
		K_{11}&K_{12}& K_{13}\\
		K_{21}&K_{22}&K_{23}\\
		K_{31}&K_{32}&K_{33}
	\end{array}\!
	\right ]\!\!, ~W\!\!=\!\!\left [\!
	\begin{array}{ccc}
		w_{11}&w_{12}& w_{13}\\
		w_{21}&w_{22}&w_{23}\\
		w_{31}&w_{32}&w_{33}
	\end{array}\!
	\right ]\!\!.
\end{aligned}
\end{equation}
In formula \eqref{e:KW},  $K_{ij}$'s  are real blur kernels,   $w_{ij}$'s belong to the interval $[0,1]$ with satisfying  $\sum_{j=1}^{3}w_{ij}=1$ for $i=1,2,3$, and the symbol $\odot$ denotes the Hadamard product.
Here,  $\star$ denotes the two-dimensional convolution operator and 
the formula \eqref{e:blurmodel0}  characterizes the blurring process  $\hat{u}_i(x,y)=\sum_{j=1}^3(K_{ij}w_{ij})\star u_j(x,y)$ for $i=1,2,3$.

Next we introduce a blurring process by using quaternion operators. 
\begin{myDef}\label{d:qblur}
Suppose $\Qq=Q_0+Q_1\iq+Q_2\jq+Q_3\kq$ is a quaternion operator, where $Q_i$'s are real blur kernels ($i=0,1,2,3$). 
A quaternion blur operator on a color image $\uq(x,y)$ 
is defined by 
 \begin{eqnarray}\label{e:qblur}
& & \!\!\!\!\! \Qq \textcircled{$\star$}\uq(x,y) \nonumber \\
&=& \!\!\!\!\!
-Q_1\star u_1(x,y)-Q_2\star u_2(x,y)-Q_3\star u_3(x,y) \nonumber \\
&& \!\!\!\!\!
+ (Q_0\star u_1(x,y)-Q_3\star u_2(x,y)+Q_2\star u_3(x,y))\iq \nonumber \\
&&\!\!\!\!\! 
+ (Q_3\star u_1(x,y)+Q_0\star u_2(x,y)-Q_1\star u_3(x,y))\jq \nonumber \\
&& \!\!\!\!\!
-(Q_2\star u_1(x,y)-Q_1\star u_2(x,y)-Q_0\star u_3(x,y))\kq. \nonumber \\
\end{eqnarray}
\end{myDef}

The  blur operator $\Qq$ depicts the rendering between color channels. We present a mathematically equivalent characterization of color image blurring, which is feasible to develop fast  algorithms.
Define 
\begin{equation}\label{e:Bmatrix}
B=\left [\begin{array}{c:ccc}
		B_{11} & B_{12} &B_{13} & B_{14}\\
		\hdashline
		B_{21} &   		&		 &		  \\
		B_{31} & 		&\Kq&			\\
		B_{41} &   		&		 &
	\end{array}\right]\!\!,
\end{equation}
where $B_{ij}$'s are arbitrary  operators of the same size of $K_{ij}$'s in \eqref{e:KW}.
Let  $Q=(B+J_4B{J_4}^T+R_4B{R_4}^T+S_4B{S_4}^T)/4$ and $R=(3B-J_4B{J_4}^T-R_4B{R_4}^T-S_4B{S_4}^T)/4$,  where  $J_4, R_4$ and $ S_4$ are unitary operators defined in \cite{jwzc18}. 

\begin{theorem}\label{t:qblur}With the above notation,  the cross-channel  blurring process  \eqref{e:blurmodel0}  is equivalent to
\begin{equation}\label{e:blurmodel0q}
	\hat{u}(x,y)=\hat{K} \star u(x,y) = \mathcal{T}( \Qq \textcircled{$\star$} \uq(x,y)+\rqq(x,y) ),
\end{equation}
where quaternion operator  $ \Qq=\Re^{-1}(Q)$ and  quaternion function  ${\rqq}(x,y)=\Re_c^{-1}(R\star \Re_c(\uq(x,y))$.
Especially, one quaternion operator satisfying \eqref{e:blurmodel0q}   is  $\Qq=Q_0+Q_1\iq+Q_2\jq+Q_3\kq$ with 
\begin{equation}\label{e:Q0123}
	\begin{aligned}
		Q_0&=\frac{1}{4}(w_{11}K_{11}+w_{22}K_{22}+w_{33}K_{33}),\\
		Q_1&=\frac{1}{4}(w_{32}K_{32}-w_{23}K_{23}),\\
		Q_2&=\frac{1}{4}(w_{13}K_{13}-w_{31}K_{31}),\\
		Q_3&=\frac{1}{4}(w_{21}K_{21}-w_{12}K_{12}),
	\end{aligned}
	\end{equation}
	where $w_{ij}$ and $K_{ij}$ are given in (\ref{e:KW}). 
\end{theorem}
\begin{proof}
The proof of Theorem \ref{t:qblur} is given in the supplementary material.
\end{proof}

The quaternion blur operator $\Qq$ can be interpreted as follows.
In the formula \eqref{e:Q0123},  the scaling factor 1/4 is the same in $Q_0$, $Q_1$, $Q_2$ and $Q_3$.
Here $Q_0$ can be viewed as the sum of the intra-blurring operators $K_{ii}$ according to 
their corresponding weights $w_{ii}$, i.e., $w_{11}K_{11}+w_{22}K_{22}+w_{33}K_{33}$. 
While $Q_1$, $Q_2$ and $Q_3$ can be viewed as the difference of the 
inter-blurring operators $w_{ij} K_{ij} - w_{ji} K_{ji}$ (or $w_{ji} K_{ji}- w_{ij} K_{ij}$) on the two different color channels ($i \ne j$). It is interesting to note that when $\hat{K}$ symmetric 
(i.e., $\hat{K}_{ij} = w_{ij} K_{ij} = w_{ji} K_{ji} = \hat{K}_{ji}$), $Q_1$,
$Q_2$ and $Q_3$ are zero matrices.
In this case, we refer to 
color blurring from the $i$-th channel to the $j$-th channel is the same as that from the $j$-th channel to the $i$-th channel.
 It is clear that ${\bf Q}$ is equal to $Q_0$, and the resulting output of each channel is equal to 
$\hat{u}_i = Q_0 u_i$ for $i=1,2,3$. Because of color blurring 
symmetry, we reformulate the whole blurring operator $\hat{K}$ into a 
simplied real form of ${\bf Q}$ such that 
each channel output is the sum of the intra-blurring operators on each channel input.
When $\hat{K}$ is not symmetric, $Q_1$, $Q_2$ and $Q_3$ are not equal to zero and therefore the quaternion blurring 
operator is not real but the color blurring effect of each imaginary component depends on the difference 
of the color blurring from the $i$-th channel to the $j$-th channel and 
the color blurring from the $j$-th channel to the $i$-th channel.

\begin{figure}[htbp]
	\centering
\subfigure[Original color image]{\includegraphics[width=0.22\textwidth,height=0.16\textwidth]{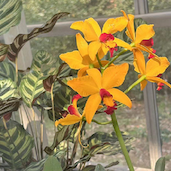}}~~~~
\subfigure[Blurred color image by  $\Kq$ as in \eqref{e:blurmodel0} and \eqref{e:KW}]{\includegraphics[width=0.22\textwidth,height=0.16\textwidth]{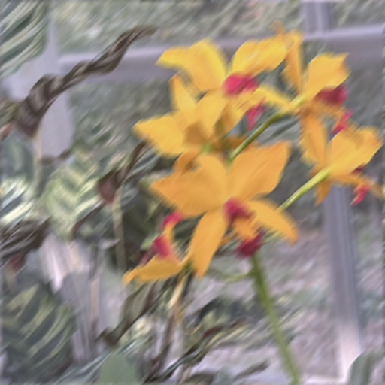}}\\
\subfigure[Blurred color image by  $\Qq$ as in \eqref{e:qblur}]{\includegraphics[width=0.22\textwidth,height=0.16\textwidth]{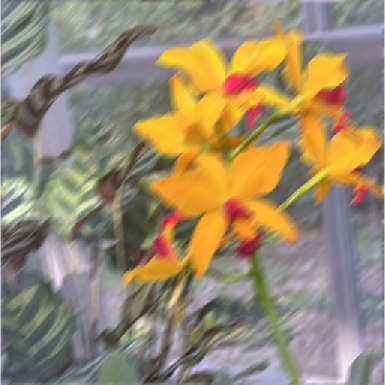}}~~~~
\subfigure[The magnitude in difference between   (b) and (c)
]{\includegraphics[width=0.22\textwidth,height=0.16\textwidth]{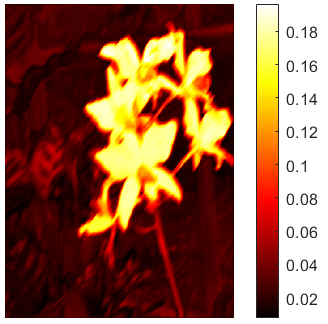}}

	\caption{A clear picture of the real-world  object (Cattleya$\times$hybrida H.J.Veitch) and the simulations of motional blur  (length=$15$, angle=$45^\circ$, symmetric)
	by using $\Kq$ and $\Qq$.
}
	\label{fig:iphoneblur2}
\end{figure}

\begin{figure}[htbp]
	\centering
\subfigure[Original color image]{\includegraphics[width=0.22\textwidth,height=0.16\textwidth]{motionblur4test/originalsmall.png}}~~~~
\subfigure[Blurred color image by  $\Kq$ as in \eqref{e:blurmodel0} and \eqref{e:KW}]{\includegraphics[width=0.22\textwidth,height=0.16\textwidth]{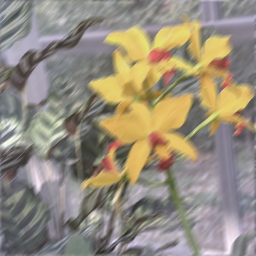}}\\
\subfigure[Blurred color image by  $\Qq$ as in \eqref{e:qblur}]{\includegraphics[width=0.22\textwidth,height=0.16\textwidth]{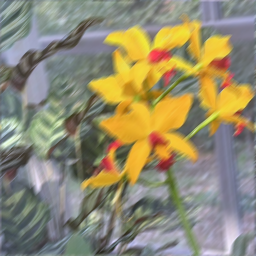}}~~~~
\subfigure[The magnitude  in difference between   (b) and (c)
]{\includegraphics[width=0.22\textwidth,height=0.16\textwidth]{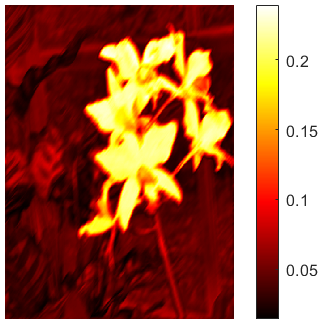}}

	\caption{A clear picture of the real-world  object (Cattleya$\times$hybrida H.J.Veitch) and the simulations of motional blur  (length=$15$, angle=$45^\circ$, asymmetric)
	by using $\Kq$ and $\Qq$.
}
	\label{fig:iphoneblur}
\end{figure}

For comparison,  we show the simulations of symmetric motional blur  by  using $\Kq$ and $\Qq$  in Figure \ref{fig:iphoneblur2} (b) and (c), respectively.
In Figure \ref{fig:iphoneblur2} (d), we show the magnitude difference between the two simulation figures.
Similarly, we show the case of asymmetric motional blur in Figure \ref{fig:iphoneblur}.

Let us turn back to construct the fidelity term of color image restoration model \eqref{e:gene}.
To reduce different kinds of noise, we define the distance between the observed and original color images by $L_p$ norm as
\begin{equation}\label{e:fidelity0}F(\uq,\zq,\Kq)=\frac{1}{p}\int_\Omega|\Kq\star \mathcal{T}(\uq)-\mathcal{T}(\zq)|^pdxdy, ~~ p \ge 1.\end{equation} 
According to Theorem \ref{t:qblur}, we have a quaternion operator form
\begin{equation}\label{e:fidelity}F(\uq,\zq,\Qq,\rqq)=\frac{1}{p}\int_\Omega|\Qq\textcircled{$\star$}\uq+\rqq-\zq|^pdxdy, ~~p \ge 1.
\end{equation} 
In formula \eqref{e:fidelity},  the quaternion convolution operator $ \Qq $ and the redundant information $\rqq$ are formed by splitting according to Theorem \ref{t:qblur}.

\subsection{Cross-space total variation regularization functional}\label{ss:regularization}
The regularization term $J(u,\lambda)$ is a functional of variations in  different color spaces of the targeted color image. It aims at extracting cross-channel features of color image, thereby enhancing the preservation of color and texture details.  
The choice of different regularization functionals is empirical and it relies on the requirement from practical applications.  
A feasible way to consolidate $J(u,\lambda)$ is to perform a linear combination of all possible regular functionals with parameters and then,  to learn the best parameters by machine learning methods.
Under the quaternion representation \eqref{e:uhatu0},  a regularization crossing HSV and RGB color spaces can be defined by 
\begin{equation}\label{e:J-HSV-RGB}J(\uq,\bm{\lambda})=\lambda_1 \SVTV(\uq)+\lambda_2\CTV(\mathcal{T}(\uq)),\end{equation}
where  $\SVTV$ is defined by \eqref{e:svtv}, $\CTV$ is defined by  \eqref{ctv1} or \eqref{ctv2}, and  $\lambda_1,\lambda_2\in\mathbb{R}$. 
$ \SVTV $ and $ \CTV $ correspond to the HSV color space, which is best performant according to the SSIM score, and the RGB color space, which is commonly used in image processing, respectively. Therefore, this CSTV regularization functional can complement each other in terms of color and texture information.
There are a number of well-known techniques for dynamically calculating  optimal parameters,  such as  the generalized cross-validation \cite{gk92,lln11}, the L-curve method \cite{h92} and the discrepancy principle \cite{wc11}.  Here, we generalize the L-curve method to a new L-surface method to determine optimal parameters simultaneously.  

\begin{figure}[htbp]
	\centering
	\subfigure[PSNR values in relation to $\lambda_1, \lambda_2$.]{\includegraphics[width=0.24\textwidth,height=0.16\textwidth]{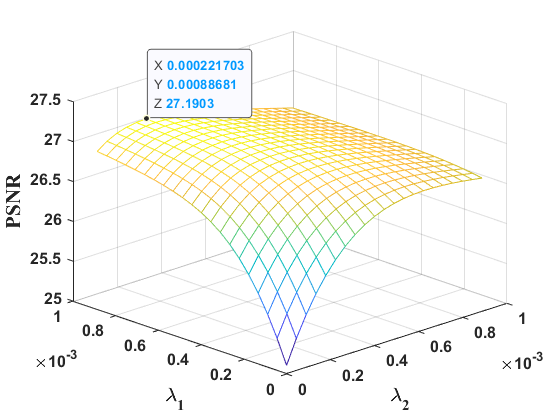}}
	\subfigure[SSIM values in relation to $\lambda_1, \lambda_2$.]{\includegraphics[width=0.24\textwidth,height=0.16\textwidth]{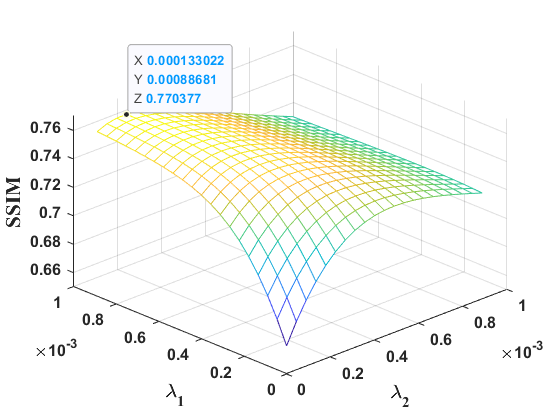}}
	\caption{The values of PSNR and SSIM for the restored image ‘statues’ in different $\lambda_1, \lambda_2$ conditions.
	}
	\label{fig:para_lambda}
\end{figure}

The L-surface method is to find a sweet spot on a surface with respect to a certain quality index of 
the parameter $\bm{\lambda}$.  
To be clear, we describe the L-surface method with step size $5\times10^{-5} $ by applying it to determine the optimal $\bm{\lambda}=(\lambda_1,\lambda_2)$ in \eqref{e:J-HSV-RGB}. Suppose we have an observed color image, say  'statues' in Figure \ref{fig:modelcom_sym},  under the setting of Example \ref{exam1}.  
The upper bound of parameters is set as $b=2\sqrt{N\sigma^2 } \times 10^{-3}$,  where $N$ is image size and $\sigma$ is the noise level.
The surface of  PSNR values of restored color images is plotted in Figure \ref{fig:para_lambda} (a) according to  $\lambda_1, \lambda_2 \in (0,b)$.  High PSNR values are obtained with $\lambda_1 \in (0.4 \times 10^{-3}, b)$ and $\lambda_2 \in (0.2 \times 10^{-3}, 0.5 \times 10^{-3})$ and  the highest PSNR value is $27.1903$ at $(\lambda_1=0.887\times 10^{-3},~ \lambda_2=0.222\times 10^{-3})$.  
The surface of  SSIM values of restored color images is plotted in Figure \ref{fig:para_lambda} (b) according to  $\lambda_1, \lambda_2 \in (0,b)$. 
 High SSIM values are obtained with $\lambda_1 \in (0.5 \times 10^{-3},b)$ and $\lambda_2 \in (0.1 \times 10^{-3}, 0.3 \times 10^{-3})$, and the highest SSIM value is $0.7704$ at $(\lambda_1=0.887\times 10^{-3},~\lambda_2=0.133\times 10^{-3})$. 
By the weighted average surface of above PSNR and SSIM surfaces,  we obtain the sweet spot $\bm{\lambda}=(0.887\times 10^{-3},~0.222\times 10^{-3})$ that leads to the optimal combination of SVTV and CTV regularization functionals.

Obviously, the color image restoration model \eqref{e:gene} with  $J(\uq,\bm{\lambda})$ as in \eqref{e:J-HSV-RGB} can obtain a broader range of pre-existing knowledge about color image than the model with $\SVTV(\uq)$ or $\CTV(\mathcal{T}(\uq))$.  In Figure \ref{fig:para_lambda}, the PSNR and SSIM values at boundary  $\lambda_1 = 0$ or $\lambda_2 = 0$ are smaller than  $27.1903$ and $0.7704$, respectively.  So one can conclude that  color image restoration models with cross-color space regularization functional  perform better than that with single color space regularization functional. 
With the introduction of cross-color space regularization, more image prior information is available to the former, enriching the ability to interpret color images.

\begin{remark}
 The SVTV regularization functional defined by \eqref{e:svtv} with \eqref{matrix:C} and \eqref{e:sv_norm} is based on the quaternion representation of color image in the RGB color space (see \cite{pech97} and \cite{sangwine96}).   It reflects the edge and texture information in the HSV color space.  
So  the CSTV regularization functional defined by \eqref{e:J-HSV-RGB}  has a quaternion variable in the RGB color space but couples the features in two color spaces. 
\end{remark}

\subsection{A new color image restoration model }
Now we  propose  a color image restoration model with a CSTV regularization term and a fidelity with quaternion blur operator.  To make a clear description, we concentrate into two color spaces: the widely used RGB color space and the more human eye-perceiving HSV color space.

For an observed image $\zq$,  
a new  CSTV regularization model for  color image restoration is
	\begin{equation}\label{m:CSmodel}
		\begin{aligned}
		\widehat{\uq}=\mathop{\arg\min}\limits_{\uq(x,y) \in \mathbb{BV}(\Omega)}~J(\uq,\bm{\lambda})+ \frac{1}{p}\int_\Omega|\Qq\textcircled{$\star$}\uq+\rqq-\zq|^p dxdy,
		\end{aligned}
	\end{equation}
where  $J(\uq,\bm{\lambda})$ is defined as in \eqref{e:J-HSV-RGB} and $p$ is a  positive integer not less than $1$.

This CSTV regularization model \eqref{m:CSmodel} contains one  fidelity term and one cross-color space regularization term. 
The quaternion representation of fidelity term  leads to many  advantages, such as the values of three color channels of a color pixel are flocked together and their physical meanings are preserved  in the whole color image processing based on quaternion computation.  
On one hand, we utilize quaternion matrices to represent color images. This approach allows for the integrated processing of the red, green, and blue channels of a color image, maintaining the proportional relationships between the channels and effectively preventing the mutual artifacts caused by splitting and recombining the channels. On the other hand, we introduce cross-space total variation regularization. Specifically, we consider the coupling of SVTV and CTV regularization terms, encompassing both the HSV and RGB color spaces. The SVTV and CTV regularization terms effectively preserve edge and texture information within their respective color spaces. The cross-space total variation regularization leverages the strengths of regularization terms from different color spaces, enhancing the breadth and depth of prior information utilized during image restoration.

 Especially, color distortion will be hugely reduced in the recovered color image by the new model.
 The cross-color space regularization functionals complement each other with color and texture information in different color spaces. In this way, the loss of information caused by considering only one color space will be decreased. 
Parameters $\lambda_1$ and $\lambda_2$ play in two roles: one is the regularization parameters to balance  regularization and fidelity terms,  and another is  the balance parameters between RGB and HSV color spaces.
When $\lambda_1=0$, the model degenerates to the classical CTV model; when $\lambda_2=0$, the model degenerates to the SVTV model; and when both $\lambda_1$ and $\lambda_2$ are greater than 0, the information of saturation, value, red, green and blue is used for color image restoration.

Next, we analyze the solvability of the minimization problem \eqref{m:CSmodel}.  
\begin{theorem}\label{t:existence}
	There is at least one solution of the CSTV regularization model \eqref{m:CSmodel}  with $p=2$ and moreover,  the solution is unique when  $\uq \mapsto\Qq\textcircled{$\star$}\uq+\rqq$ is injective.
\end{theorem}
\begin{IEEEproof}
If $\uq(x, y)$ is a constant value function, then $\SVTV(\uq)=0$  and $\CTV(\mathcal{T}(\uq))=0$. The energy in \eqref{m:CSmodel} thus becomes finite. This implies that the infimum of the energy must be finite. Suppose that $\left\{\uq^{(n)}(x, y)\right\}$ is a minimizing sequence of problem \eqref{m:CSmodel}. Consequently, a constant $M>0$ exists such that
	$$\SVTV(\uq^{(n)})\leq M.$$
	We get that $\left\{\SVTV(\uq^{(n)})+\sum\limits_{i=1}^3\|u_i^{(n)}\|_{L^1}\right\}$ is uniformly bounded by combining the boundedness of $u_1^{(n)},u_2^{(n)},u_3^{(n)}$. Noting the compactness referenced in \cite{jnw19}, we know that there exist $u_1^{\min},u_2^{\min},u_3^{\min}$ such that
	\begin{equation*}
		\begin{aligned}
			u_i^{(n)}(x, y)\!\xrightarrow{{L^1}(\Omega)}\!u_i^{\min}(x, y),~~~u_i^{(n)}(x, y)\!\rightarrow\! u_i^{\min}(x, y)\\
			a.e.~(x, y) \!\in\! \Omega,~~i=1,2,3.
		\end{aligned}
	\end{equation*}
Recall that $\mathcal{T}(\Qq\textcircled{$\star$}\uq(x,y)+\rqq(x,y))=\Kq\star \mathcal{T}(\uq(x,y))$.
	Subsequently, the convergence of the following results is valid:
	\begin{equation*}
		\begin{aligned}
					\|\Kq\star \mathcal{T}(\uq^{(n)}(x,y))-\mathcal{T}(\zq(x,y))\|^2\rightarrow ~~~~~~~~~~~~~~~~~~~~~~~~\\
					\|\Kq\star \mathcal{T}(\uq^{\min}(x,y))-\mathcal{T}(\zq(x,y))\|^2~~
			a.e.~~(x, y) \in \Omega.
		\end{aligned}
	\end{equation*}
	Taking advantage of the Fatou's lemma, one has
	\begin{equation*}
		\begin{aligned}
				\lim\inf\int_\Omega\|\Kq\star  \mathcal{T}(\uq^{(n)}(x,y))- \mathcal{T}(\zq(x,y))\|^2dxdy\\
			\geq \int_\Omega\|\Kq\star  \mathcal{T}(\uq^{\min}(x,y))- \mathcal{T}(\zq(x,y))\|^2dxdy.
		\end{aligned}
	\end{equation*}
	On account of the lower semi-continuity of $\CTV(\mathcal{T}(\uq))$ and $ \SVTV(\uq)$,
	\begin{equation*}
		\begin{aligned}
			&\lim\inf~\CTV(\mathcal{T}(\uq^{(n)}))\geq \CTV(\mathcal{T}(\uq^{\min})),\\
			&\lim\inf~\SVTV(\uq^{(n)})\geq \SVTV(\uq^{\min}),
		\end{aligned}
	\end{equation*}
	then we obtain
	\begin{equation*}
		\begin{aligned}
					&\lim\inf ~(\lambda_1\SVTV(\uq^{(n)})+\lambda_2\CTV(\mathcal{T}(\uq^{(n)}))\\
			&~~~~~~~+\int_\Omega\|\Kq\star u^{(n)}(x,y)-z(x,y)\|^2dxdy) \\
			&\geq (\lambda_1\SVTV(\uq^{\min})+\lambda_2\CTV(\mathcal{T}(\uq^{\min}))\\
			&~~~~~~+\int_\Omega\|\Kq\star (\mathcal{T}(\uq^{\min}(x,y))-(\mathcal{T}(\zq(x,y))\|^2dxdy).
		\end{aligned}
	\end{equation*}
	It leads to the existence of the solution of \eqref{m:CSmodel} in the main body. If $\uq \mapsto \Qq\textcircled{$\star$}\uq+\rqq$ $(=\Kq\star(\mathcal{T}(\uq))$ is injective, the uniqueness of the solution can be obtained quickly by combining the convexity of SVTV$(\uq)$ and CTV$((\mathcal{T}(\uq))$.
\end{IEEEproof}

An important advantage of the  CSTV regularization model \eqref{m:CSmodel} is that it takes into account the color channels coupling and the local smoothing within each channel. This implicitly leads to achieving better results in terms of both image texture and color fidelity. One can get the explanation from  the corresponding Euler-Lagrange (EL) equation. 
Let  $\Kq^T$ denote the conjugate transpose of $\Kq$ and define $u=\mathcal{T}(\uq)$.  The EL equations of \eqref{m:CSmodel} is 
\begin{equation*}
	\begin{aligned}[c]
		&\frac{\lambda_1}{3} \nabla\cdot \Bigg( \frac{\nabla\left(Cu  \right) }{\sqrt{(|\partial_x\uq|_s^2)+(|\partial_y\uq|_s^2)}}+\alpha\frac{\nabla u  }{\sqrt{(|\partial_x\uq|_v^2)+(|\partial_y\uq|_v^2)}} \Bigg)\\
		&\qquad +\lambda_2\nabla \cdot\frac{\nabla u}{\|\nabla u\|_2}
		-\left( \Kq\star u-\mathcal{T}(\zq) \right) \star\Kq^T=0.	
	\end{aligned}
\end{equation*} 
An interesting finding is that the first two denominators can be further simplified and  the gradient of the value component is consistent with that of  CTV. 
The EL equation of \eqref{m:CSmodel} becomes
\begin{equation}\label{e:EL_whole}
	\begin{aligned}
		&\nabla \cdot\left(\lambda_1\frac{\nabla (Cu) }{\|\nabla (Cu)\|_2}+(\frac{\sqrt{3}}{3}\alpha\lambda_1+\lambda_2)\frac{\nabla u}{\|\nabla u\|_2}\right)\\
		&\qquad- (\Kq\star u-\mathcal{T}(\zq)) \star\Kq^T=0.
	\end{aligned}
\end{equation}
The EL equation \eqref{e:EL_whole} indicates that the proposed model  \eqref{m:CSmodel} takes into account both the diffusion coefficients of the channel coupling over the saturation and value components and the diffusion coefficients of the summation of the individual channels over the RGB color space.

To end this section, we present the dual form of  model \eqref{m:CSmodel}
	\begin{equation}\label{m:dualCSmodel}
		\min\limits_{\uq}\!\max\limits_{\|g\|\leq 1
			\atop
			\|h\|\leq 1}\lambda_1\!\left<\diag(1,1,\alpha)P\mathcal{T}(\uq),\nabla g\right>+
 \atop
\qquad\qquad\qquad~~ \lambda_2\!\left<\mathcal{T}(\uq),\nabla h\right>+ \frac{1}{2}\|\Qq\textcircled{$\star$}\uq+\rqq-\zq\|^2,
	\end{equation}
	where $$
		P=\left [\begin{array}{ccc}
			\frac{1}{\sqrt{2}}&\frac{-1}{\sqrt{2}}& 0\\
			\frac{1}{\sqrt{6}}&\frac{1}{\sqrt{6}}&\frac{-2}{\sqrt{6}}\\
			\frac{1}{\sqrt{3}}&\frac{1}{\sqrt{3}}&\frac{1}{\sqrt{3}}
		\end{array}\right]\!\!.$$
The solution $\uq(x,y)$ of \eqref{m:dualCSmodel} can be differentiable or not, which expands the feasible solution set.   
The detailed derivation of the  EL equation  \eqref{e:EL_whole}  and the dual form \eqref{m:dualCSmodel} is presented  in the supplementary material.

\section{Fast and stable algorithm}\label{s:numalg}
In this section, we present a new   fast and stable algorithm for model \eqref{m:CSmodel} based on quaternion operator splitting.

 With introducing two auxiliary variables $\wq$ and $\vq$, we  obtain an  mathematically equivalent model  of   \eqref{m:CSmodel}
\begin{equation*}
\begin{aligned}
		&\widehat{\uq}=\mathop{\arg\min}\limits_{\uq}~\lambda_1\SVTV(\wq)+\lambda_2\CTV((\mathcal{T}(\vq))+\atop	
\frac{1}{2}\|\Qq\textcircled{$\star$}\uq+\rqq-\zq\|^2,\\
		&\qquad \text{s.t.}~~\wq=\uq,~~\vq=\uq.
\end{aligned}
	\end{equation*}
The  above constrained optimal  minimization problem can be converted into an unconstrained problem:
\begin{equation}\label{p:model}
		\!\!\!\!\!\!\!\!\!\!\!\! \widehat{\uq}=\mathop{\arg\min}\limits_{\uq,\wq,\vq}~\lambda_1\SVTV(\wq)+\frac{\alpha_1}{2}\|\wq-\uq\|^2+\atop
		 	\! \frac{1}{2}\|\Qq\textcircled{$\star$}\uq+\rqq-\zq\|^2+\lambda_2\CTV((\mathcal{T}(\vq))+\frac{\alpha_2}{2}\|\vq-\uq\|^2,
\end{equation}
where $\alpha_1$ and  $\alpha_2$ are two positive parameters. 
Three unknown variables $\uq$, $\wq$ and $\vq$ can be separated into two independent groups and under the framework of alternating minimization method,  problem \eqref{p:model} can be solved by alternatively solving  two subproblems: 

\begin{itemize}

\item[$\bullet$]$\uq$-subproblem
	
	With fixing $\wq$ and $\vq$,  the minimization problem \eqref{p:model} is equivalently reduced to
	\begin{equation}\label{model_u}
		\mathop{\arg\min}\limits_{\uq}~	\frac{\alpha_1}{2}\|\uq-\wq\|^2+ \frac{\alpha_2}{2}\|\uq-\vq\|^2+\atop \frac{1}{2}\|\Qq\textcircled{$\star$}\uq+\rqq-\zq\|^2.
	\end{equation}

\item[$\bullet$]$(\wq,\vq)$-subproblem
	
	With fixing $\uq$,   the minimization problem \eqref{p:model} is equivalently reduced to
	\begin{equation}\label{model_wv}
		\mathop{\arg\min}\limits_{\wq,\vq}~\lambda_1\SVTV(\wq)+\frac{\alpha_1}{2}\|\wq-\uq\|^2+\atop \lambda_2\CTV((\mathcal{T}(\vq))+ \frac{\alpha_2}{2}\|\vq-\uq\|^2.
	\end{equation}

\end{itemize}

\subsection{Quaternion operator splitting method for  $\uq$-subproblem} \label{ss:uqsubproblem}
Now we consider the $\uq$-subproblem \eqref{model_u} and use its real representation to introduce a quaternion operator splitting method.  

From the analysis in Section \ref{ss:fidelity}, the real representation  of \eqref{model_u}  is 
	\begin{equation}\label{model_u-real}
		\mathop{\arg\min}\limits_{\uq}~	\frac{\alpha_1}{2}\|\uq-\wq\|^2+ \frac{\alpha_2}{2}\|\uq-\vq\|^2+\atop \frac{1}{2}\|B\Re_c(\uq)-\Re_c(\zq)\|^2,
	\end{equation}
where $B$ is defined by \eqref{e:Bmatrix}.
Its normal equation  is 
	\begin{equation}\label{e:normequation}
		[B^TB+(\alpha_1+\alpha_2)\Iq]\Re_c(\uq)=B^T\Re_c(\zq)+\alpha_1 \Re_c(\wq)+\alpha_2\Re_c(\vq).
	\end{equation} 
Let $Q=\Re(\Qq)$ denote the real counterpart of  quaternion blur operator $\Qq$.   The operator $B$ can be splitted into $Q+R$  with setting $R=B-Q$. 
This implies  $B^TB=Q^TQ+Q^TR+R^TR+R^TQ$. 
An iterative format of \eqref{e:normequation} is derived as
\begin{equation}\label{itls_u-real}
		\begin{aligned}
		\!\![Q^TQ+(\alpha_1+\alpha_2)\Iq]\Re_c(\uq^{k+1})\!=\!-(Q^TR+R^TQ+~~~~~~~~\\
		R^TR)\Re_c(\uq^{k})+(Q+R)^T\Re_c(\zq)+\alpha_1\Re_c(\wq)+\alpha_2\Re_c(\vq).
		\end{aligned}
	\end{equation}

Let   $\Aq=\Re^{-1}(Q^TQ+(\alpha_1+\alpha_2)\Iq)$, $\xq=\uq^{k+1}$ and $\bq=\Re_c^{-1}(-(Q^TR+R^TQ+R^TR)\Re_c(\uq^{k})+(Q+R)^T\Re_c(\zq)+\alpha_1\Re_c(\wq)+\alpha_2\Re_c(\vq))$. 
Then we can construct an equivalent quaternion linear system of \eqref{itls_u-real},
	\begin{equation}\label{itls_u-quaterion}
		\begin{aligned}
		\Aq\xq=\bq.
		\end{aligned}
	\end{equation}
So the solution of the $\uq$-subproblem \eqref{model_u}  can be computed by iteratively solving \eqref{itls_u-quaterion}. 
That means we have proposed a quaternion operator splitting method for solving the $\uq$-subproblem, where the coefficient  matrix is splitted into a quaternion operator and a residual operator.   This leads to an advantage that the color information of restored color image $\uq$ is preserved by quaternion algebra operations in the solving process.  

The core work becomes solving \eqref{itls_u-quaterion}.  QGMRES \cite{jing21} is  feasible to solve  this  quaternion linear system.	However,  we can further apply the Hermitian and positive definite  properties of $\Aq$ to develop a new quaternion conjugate gradient method.   Its structure-preserving version is given in Algorithm \ref{code:cgQ}.  The quaternion matrix-vector product  is implemented by real operation, i.e.,  $\Aq\xq=\Re_c^{-1}(\Re(\Aq)\Re_c(\xq))$.

\begin{algorithm}[htbp]
	\setcounter{algorithm}{0}
	\caption{Quaternion Conjugate Gradient Method}
	\label{code:cgQ}
	\begin{algorithmic}[1]
		\STATE \textbf{Initialization}  Set the stopping criteria $tol>0$ and the initial solution  $\xq_0=0$.
	          \STATE $k=0$, $\rqq_k=\bq-\Re_c^{-1}(\Re(\Aq)\Re_c(\xq_k))$ and let  $\pq_k=\rqq_k$.
		\WHILE {$\|\Re_c(\rqq_k)\|_2>tol$}
                   \STATE $\qq_k=\Re_c^{-1}(\Re(\Aq)\Re_c(\pq_k))$,
		\STATE Compute $a_k=\frac{\left<\rqq_k,\rqq_k\right>}{\left<\pq_k,\qq_k\right>}$,
	         \STATE Compute $\xq_{k+1}=\xq_k+a_k\pq_k$,
		\STATE Compute $\rqq_{k+1}=\rqq_k+a_k\qq_k$,
		\STATE Compute  $b_k=\frac{\left<\rqq_{k+1},\rqq_{k+1}\right>}{\left<\rqq_k,\rqq_k\right>}$,
	          \STATE Compute $\pq_{k+1}=\rqq_{k+1}+b_k\pq_k$,
		\STATE $k=k+1$.
		\ENDWHILE
	\end{algorithmic}
\end{algorithm}

\begin{remark}
	Since $ Q^TQ $ is symmetric and semi-positive definite, the coefficient matrix $ [Q^TQ+(\alpha_1+\alpha_2)\Iq] $ is surely a symmetric positive definite matrix. These conditions ensure that each inner iteration in solving $\xq$ in Algorithm \ref{code:cgQ} effectively reduces the residuals and moves towards a numerical solution. In other words, the convergence and effectiveness of the iterative solution of the $ \uq $-subproblem are guaranteed.
\end{remark}

\subsection{Augmented Lagrangian method for $(\wq,\vq)$-subproblem} \label{ss:wqvqsubproblem}
Now, we present a new augmented Lagrangian method for solving $(\wq,\vq)$-subproblem \eqref{model_wv}.  Since the two variables are independent of each other and have no intersecting terms, we construct the augmented Lagrangian schemes of computing $\wq$ and $\vq$, respectively.

\begin{itemize}	
	\item[$\bullet$]$\wq$-subproblem
	
	With $\uq$ and $\vq$ fixed,   the minimization problem \eqref{model_wv} is reduced to
	\begin{equation}\label{model_w}
		\mathop{\arg\min}\limits_{\wq}~\lambda_1\SVTV(\wq)+\frac{\alpha_1}{2}\|\wq-\uq\|^2.
	\end{equation}

	Define 
$(\Dq_x\wq)_{i,j}=\wq(i,j)-\wq(i-1,j)$, $(\Dq_y\wq)_{i,j}=\wq(i,j)-\wq(i,j-1)$, 
$\sq=P\mathcal{T}(\wq)$,  and $\qq=P\mathcal{T}(\uq)$.
By introducing  auxiliary variables $t_k^x=D_xs_k$  and  $t_k^y=D_ys_k$, we obtain 
	the augmented Lagrangian  equation of  the $\wq$-subproblem  \eqref{model_w}, 
	\begin{equation}\label{alm}
		\begin{aligned}
			&\!\!\!\!\frac{\lambda_1}{\alpha_1}\sum\limits_{i=1}^m\sum\limits_{j=1}^n\left(\sqrt{\sum\limits_{k=1}^2|(t_k^x)_{ij}|^2+|(t_k^y)_{ij}|^2}\right.\\
			&\left.+\alpha\sqrt{|(t_3^x)_{ij}|^2+|(t_3^y)_{ij}^2|}\right)+\frac{1}{2}\|\sq-\qq\|^2\\
			&+\sum\limits_{i=1}^3[(\tau_i^x,t_i^x-\Dq_xs_i)+(\tau_i^y,t_i^y-\Dq_ys_i)]\\
			&+\frac{\beta}{2}\sum\limits_{i=1}^3(\|t_i^x-\Dq_xs_i\|^2+\|t_i^y-\Dq_ys_i\|^2),
		\end{aligned}
	\end{equation}
where  $\tau_i^x$ and $\tau_i^y$ are  Lagrangian multipliers and  $\beta$ is a positive penalty parameter.

	\item [$\bullet$]$\vq$-subproblem
	
	With  $\uq$ and $\wq$ fixed,   the minimization problem \eqref{model_wv} is reduced to
	\begin{equation}\label{model_v}
		\mathop{\arg\min}\limits_{\vq}~\lambda_2\CTV(\mathcal{T}(\vq))+\frac{\alpha_2}{2}\|\vq-\uq\|^2.
	\end{equation}
By introducing  auxiliary variables $l_k^x=D_xv_k$  and  $l_k^y=D_yv_k$, we obtain 
	the augmented Lagrangian  equation of  the $\vq$-subproblem  \eqref{model_v}, 
\begin{equation}\label{alm_v}
	\begin{aligned}		&\!\!\!\!\frac{\lambda_2}{\alpha_2}\sum\limits_{i=1}^m\sum\limits_{j=1}^n\sqrt{\sum\limits_{k=1}^3|(l_k^x)_{ij}|^2+|(l_k^y)_{ij}|^2}+\frac{1}{2}\|\vq-\uq\|^2\\ &+\sum\limits_{i=1}^3((\eta_i^x,l_i^x-\Dq_xv_i)+(\eta_i^y,l_i^y-\Dq_yv_i))\\ &+\frac{\beta_2}{2}\sum\limits_{i=1}^3(\|l_i^x-\Dq_xv_i\|^2+\|l_i^y-\Dq_yv_i\|^2).
	\end{aligned}
\end{equation}
where  $\eta_i^x$ and $\eta_i^y$ are  Lagrangian multipliers and  $\beta_2$ is a positive penalty parameter.
	
\end{itemize}

The solutions of \eqref{model_w} and \eqref{model_v} can be computed by many well-known methods; see \cite{jnw19,brc08} for instance.
Overall, the $(\wq,\vq)$-subproblem needs to compute several soft shrinkage functions and solve two real linear systems. 
According to the classic framework of the ADMM \cite{e09,jd92}, we construct a method to solve  \eqref{alm} and \eqref{alm_v} within three steps  in the supplementary material.

\subsection{A new fast and stable algorithm based on quaternion operator splitting}
Now we present a new algorithm  for model \eqref{m:CSmodel} based on the methods  in Sections \ref{ss:uqsubproblem} and \ref{ss:wqvqsubproblem}.  The puedo code is given  in Algorithm \ref{code:Qcstv}.

\begin{algorithm}[htbp]
	\caption{Quaternion Operator Splitting Method}
	\label{code:Qcstv}
	\begin{algorithmic}[1]
		\STATE \textbf{Initialization} Choose parameters $\lambda_1, \lambda_2, \alpha_1, \alpha_2$, $\tau_i^x, \tau_i^y$ and $\eta_i^x, \eta_i^y$ $(i=1,2,3), \alpha, \beta$. Set the stopping criteria $tol>0$ and $\uq^{0}=\bf{0}$, $\wq^{0}=\zq, \vq^{0}=\zq, err=1, err_u=1, l=0$.
		\WHILE {$err>tol$}
		\STATE $k=0, \uq^{k}=\uq^{l}$
		\WHILE{$err_u>tol$}
		\STATE Solve $\uq^{k+1}$ by \eqref{itls_u-quaterion} using Algorithm \ref{code:cgQ},
		\STATE $k=k+1$,
		\STATE Compute $err_{u}=\frac{\|\uq^{k-1}-\uq^{k}\|_2^2}{\|\uq^{k-1}\|_2^2}$,
		\ENDWHILE
		\STATE $\uq^{l+1}=\uq^{k}$,
		\STATE Update $\wq^{l+1}$ by solving \eqref{model_w},
		\STATE Update $\vq^{l+1}$ by solving \eqref{model_v},
		\STATE $l=l+1$,
		\STATE Compute $err=\frac{\|\vq^{l-1}-\vq^{l}\|_2^2}{\|\vq^{l-1}\|_2^2}$.
		\ENDWHILE
	\end{algorithmic}
\end{algorithm}

The main idea of Algorithm \ref{code:Qcstv} is to alternately solve \eqref{model_u} and \eqref{model_wv} with proper initial values.  The only problem left for discussion is how to determine their augmented Lagrange  parameters $\alpha_i$'s. This is one of the well-known parameter-selection problems that stand for a long history.  
Here, we apply the newly proposed L-surface method  in Section \ref{ss:regularization} to determine the optimal  $\alpha_i$'s. These parameters are determined using a few typical images before running before running Algorithm \ref{code:Qcstv} and, once determined, are used to process all images of the same degradation type.

\begin{figure}[htbp]
	\centering
	\includegraphics[width=0.4\textwidth,height=0.28\textwidth]{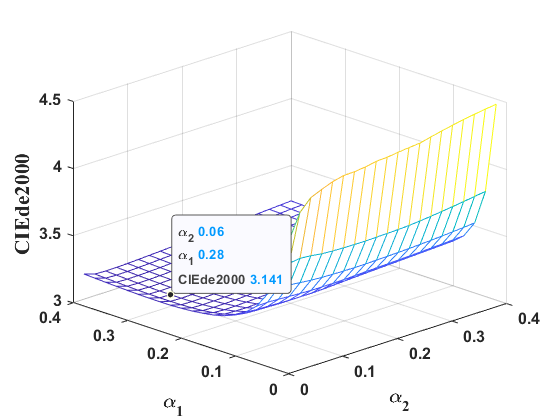}
	\caption{The values of  CIEde2000 for the restored image ‘statues’ in different $\alpha_1$ and $\alpha_2$ conditions.
	}
	\label{fig:paraset}
\end{figure}

Now we demonstrate the  parameter selection  process of $\alpha_i$'s by the L-surface method with step size $2\times10^{-2} $  in Section \ref{ss:regularization}.
Suppose that we have obtained an observed color image, say  'statues' in Figure \ref{fig:modelcom_sym},  under the setting of Example \ref{exam1}.  In case of taking CIEde2000 value  as a criterion, we plot the surface of  CIEde2000 values according to  $\alpha_1, \alpha_2 \in [0,0.4]$  of restored color images  in Figure \ref{fig:paraset}.   The optimal parameter is $\alpha_1=0.28,~\alpha_2=0.06$ at which the minimum CIEde2000 value reaches $3.1410$. 

 We have proposed a fast and stable algorithm (Algorithm \ref{code:Qcstv}) to solve the $\CSTV$ regularization model for cross-channel deblurring problem. This idea of quaternion characterization of blurring and quaternion operator splitting  can also be applied to improve the well-known color image restoration models such as $\CTV$ and $\SVTV$.  The numerical comparison will be given in Section \ref{exam3}.

\section{Numerical Experiments}\label{s:experiments}
In this section, we consider various types of cross-channel blur kernels and weight matrices to illustrate the validity of the proposed model and algorithms. 
We compare the proposed  CSTV regularization model and Algorithm \ref{code:Qcstv}  (referred to as CSTV) with  $\CTV_1$ \cite{bc98}, $\CTV_2$ \cite{brc08},  $\SVTV$ \cite{jnw19}, 
the split-algorithm-based $\CTV$ ($\rm{\CTV_{split}}$) \cite{wmh08}, $\MTV$ \cite{yyzw09}, by applying them to solve cross-channel deblurring problems.
As in \cite{jnw19}, we choose the optimal values of  the regularization parameters  in terms of PSNR for all the compared methods.
All experiments were implemented by MATLAB (R2020a) on a computer with Intel(R) Xeon(R) CPU E5-2630 @ 2.40Ghz/32.00 GB. The stopping criterion for all these iterative methods is that the norm of the successive iterations is less than $1.0\times10^{-5}$.

\begin{figure}[htbp]
	\centering
	\includegraphics[width=0.48\textwidth,height=0.22\textwidth]{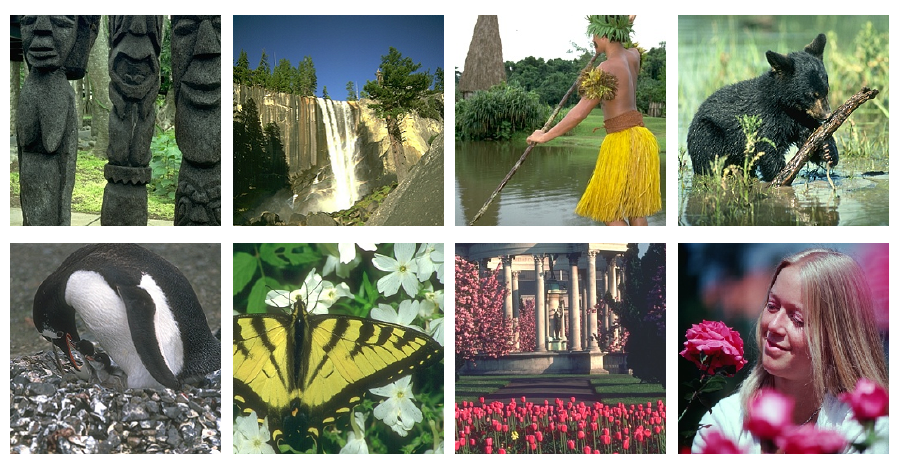}
	\caption{Ground truth images.
	}
	\label{fig:Ground-truth images}
\end{figure}

\begin{figure}[!t]
		\centering
		\includegraphics[width=0.48\textwidth]{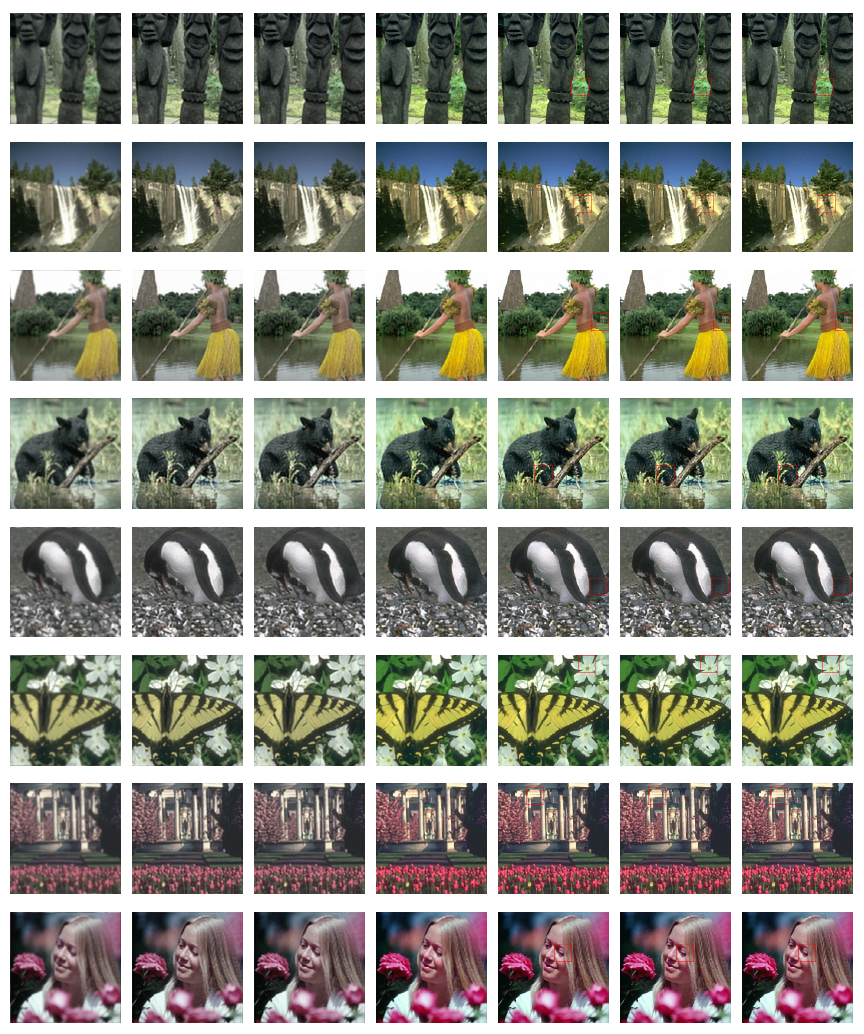}
		\caption{ Visual comparison of six methods in symmetric cross-channel deblurring. 
		From left to right: Observed images and restorations by  $\CTV_1$, $\CTV_2$,  
$\rm{\CTV_{split}}$, $\MTV$, $\SVTV$ and CSTV. (Color images have been reduced to save space. The same applies hereinafter.)
		}
		\label{fig:modelcom_sym}
	\end{figure}

We use the ground truth images of size $n\times n$ shown in Figure \ref{fig:Ground-truth images} with $n=512$. 
Let $\Uq$ and $\hat{\Uq}$  denote the original and restored color images, respectively.
The quality of the restored color image is indicated by the four standard criteria: \texttt{PSNR}, \texttt{SSIM}, \texttt{MSE}, and \texttt{CIEde2000}.
 \texttt{PSNR} means the peak signal-to-noise ratio value of $\hat{\Uq}$, defined by 
$$\texttt{PSNR}(\hat{\Uq},\Uq)=10*\log_{10}\left(\frac{255^2n^2}{\|\hat{\Uq}-\Uq\|_F^2}\right).$$
\texttt{SSIM} denotes the structural similarity  index \cite{Wbs04} of $\hat{\Uq}$ and $\Uq$, defined by
$$\texttt{SSIM}(\hat{\Uq},\Uq)=\frac{(4\mu_x\mu_y+c_1)(2\sigma_{xy}+c_2)}{(\mu_x^2+\mu_y^2+c_1)(\sigma_x^2+\sigma_y^2+c_2)},$$
where $x$ and $y$ respectively stand for the vector forms of $\hat{\Uq}$ and $\Uq$,   $c_{1,2}$ are two constants, $\mu_{x,y}$ signify the averages of $x$ and $y$, and $\sigma_{x,y}^2$ stand for the variances of $x$ and $y$, and $\sigma_{xy}$ denotes the covariance between $x$ and $y$.
\texttt{MSE} is the mean square error  value of $\hat{\Uq}$, defined by
$$\texttt{MSE}(\Uq_m,\Uq)=\frac{\|\hat{\Uq}-\Uq\|_F^2}{n^2}.$$
The \texttt{CIEde2000} color difference formula is described in \cite{Gwe05}, which is used to evaluate the color difference between the original and the restored image.

	\begin{figure}[!t]
		\centering
		\includegraphics[width=0.48\textwidth]{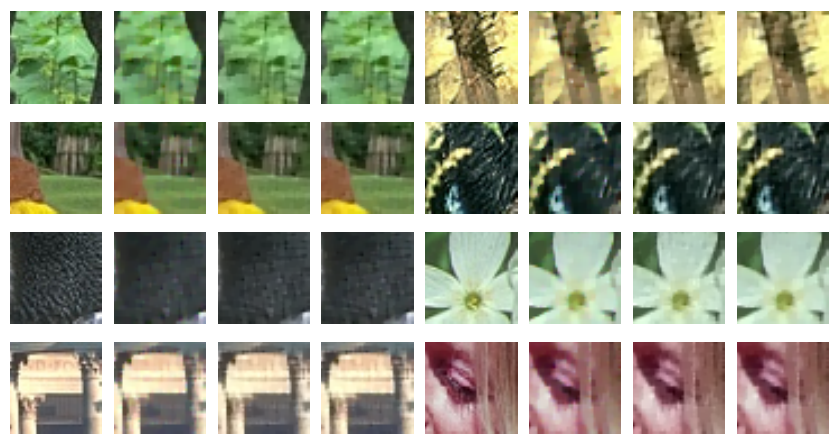}
		\caption{Corresponding zooming parts of original image and restorations by  $\MTV$, $\SVTV$ and CSTV given in Figure \ref{fig:modelcom_sym}.
		}
		\label{zoom_sym}
	\end{figure}

	\begin{table*}[htbp]
		\begin{lrbox}{\tablebox}
			\begin{tabular}{|l|c|c|c|c|c||l|c|c|c|c|c|}
				\hline
				\makecell[c]{Observed image\\(PSNR~SSIM)}
				&Method &{\tt PSNR} & {\tt SSIM} & \makecell[c]{{\tt MSE} \\(\num{1e-3})} & {\tt CIEde2000}&\makecell[c]{Observed image\\(PSNR~SSIM)}
				&Method &{\tt PSNR} & {\tt SSIM} & \makecell[c]{{\tt MSE} \\(\num{1e-3})} & {\tt CIEde2000}\\  \hline
				
				\makecell[c]{statues\\(22.6920~0.5002)}
&\makecell[c]{$\CTV_1$\\$\CTV_2$\\$\rm{\CTV_{split}}$\\$\MTV$\\$\SVTV$\\ $\CSTV$}
&\makecell[c]{24.7890\\24.7416\\26.1593\\26.9277\\\underline{27.3926}\\$\mathbf{27.4356}$}
&\makecell[c]{0.6817\\0.6740\\0.6983\\0.7374\\\underline{0.7688}\\$\mathbf{0.7718}$}
&\makecell[c]{3.3197\\3.3561\\2.4214\\2.0288\\\underline{1.8228}\\$\mathbf{1.8048}$}
&\makecell[c]{6.3750\\6.1065\\3.9703\\4.0883\\$\mathbf{3.1326}$\\\underline{3.1611}}&
			
				\makecell[c]{waterfall\\(21.3554~0.5654)}
&\makecell[c]{$\CTV_1$\\$\CTV_2$\\$\rm{\CTV_{split}}$\\$\MTV$\\$\SVTV$\\ $\CSTV$}
&\makecell[c]{22.9317\\22.8933\\25.2652\\26.3011\\\underline{26.6554}\\$\mathbf{26.7418}$}
&\makecell[c]{0.7027\\0.7097\\0.7541\\0.8008\\\underline{0.8156}\\$\mathbf{0.8169}$}
&\makecell[c]{5.0914\\5.1365\\2.9749\\2.3426\\\underline{2.1601}\\$\mathbf{2.1175}$}
&\makecell[c]{7.5840\\7.4116\\3.7826\\3.7302\\\underline{3.2863}\\$\mathbf{3.1879}$}\\
				\hline

				\makecell[c]{aborigine\\(21.1179~0.5610)}
&\makecell[c]{$\CTV_1$\\$\CTV_2$\\$\rm{\CTV_{split}}$\\$\MTV$\\$\SVTV$\\ $\CSTV$}
&\makecell[c]{22.3146\\22.29382\\26.8057\\27.7981\\\underline{28.1841}\\$\mathbf{28.2283}$}
&\makecell[c]{0.7152\\0.7181\\0.7673\\0.8107\\\underline{0.8285}\\$\mathbf{0.8293}$}
&\makecell[c]{5.8687\\5.8968\\2.0866\\1.6603\\\underline{1.5191}\\$\mathbf{1.5037}$}
&\makecell[c]{7.2164\\7.0597\\3.3089\\3.3228\\\underline{2.7941}\\$\mathbf{2.7419}$}&

				\makecell[c]{bear\\(20.2929~0.5507)}
&\makecell[c]{$\CTV_1$\\$\CTV_2$\\$\rm{\CTV_{split}}$\\$\MTV$\\$\SVTV$\\ $\CSTV$}
&\makecell[c]{21.9875\\21.9315\\23.3130\\24.3554\\\underline{24.7129}\\$\mathbf{24.8641}$}
&\makecell[c]{0.7154\\0.7134\\0.7477\\0.7907\\\underline{0.8108}\\$\mathbf{0.8111}$}
&\makecell[c]{6.3278\\6.4098\\4.6634\\3.6683\\\underline{3.3784}\\$\mathbf{3.2628}$}
&\makecell[c]{8.1320\\8.0203\\4.3142\\4.3680\\\underline{3.8824}\\$\mathbf{3.7156}$}\\
				\hline
				
				\makecell[c]{penguin\\(21.8428~0.6083)}
&\makecell[c]{$\CTV_1$\\$\CTV_2$\\$\rm{\CTV_{split}}$\\$\MTV$\\$\SVTV$\\ $\CSTV$}
&\makecell[c]{24.1892\\24.1561\\25.3151\\26.5423\\\underline{26.8982}\\$\mathbf{26.9406}$}
&\makecell[c]{0.7349\\0.7389\\0.7760\\0.8079\\\underline{0.8206}\\$\mathbf{0.8209}$}
&\makecell[c]{3.8113\\3.8405\\2.9410\\2.2170\\\underline{2.0426}\\$\mathbf{2.0228}$}
&\makecell[c]{5.7218\\5.3536\\4.5552\\4.7076\\$\mathbf{3.4735}$\\\underline{3.5037}}&
				
				\makecell[c]{butterfly\\(20.9821~0.6702)}
&\makecell[c]{$\CTV_1$\\$\CTV_2$\\$\rm{\CTV_{split}}$\\$\MTV$\\$\SVTV$\\ $\CSTV$}
&\makecell[c]{22.6934\\22.6994\\26.8852\\28.3685\\\underline{28.3065}\\$\mathbf{28.4958}$}
&\makecell[c]{0.7702\\0.7789\\0.8355\\0.8672\\\underline{0.8637}\\$\mathbf{0.8734}$}
&\makecell[c]{5.3785\\5.3710\\2.0487\\\underline{1.4560}\\1.4769\\$\mathbf{1.4139}$}
&\makecell[c]{7.8815\\7.7457\\3.0992\\3.0369\\\underline{2.8690}\\$\mathbf{2.5785}$}\\
				\hline
				
				\makecell[c]{garden\\(19.4933~0.5176)}
&\makecell[c]{$\CTV_1$\\$\CTV_2$\\$\rm{\CTV_{split}}$\\$\MTV$\\$\SVTV$\\ $\CSTV$}
&\makecell[c]{21.6959\\21.6205\\23.7840\\25.2706\\\underline{25.4766}\\$\mathbf{25.6999}$}
&\makecell[c]{0.7188\\0.7167\\0.7741\\0.8354\\\underline{0.8478}\\$\mathbf{0.8494}$}
&\makecell[c]{6.7673\\6.8857\\4.1841\\2.9712\\\underline{2.8336}\\$\mathbf{2.6916}$}
&\makecell[c]{7.7463\\7.5397\\4.9869\\4.9658\\\underline{4.6447}\\$\mathbf{4.1641}$}&
				
				\makecell[c]{girl\\(22.5950~0.6946)}
&\makecell[c]{$\CTV_1$\\$\CTV_2$\\$\rm{\CTV_{split}}$\\$\MTV$\\$\SVTV$\\ $\CSTV$}
&\makecell[c]{24.0889\\24.0870\\28.5861\\29.5834\\\underline{29.7902}\\$\mathbf{29.8912}$}
&\makecell[c]{0.7844\\0.7941\\0.8430\\0.8634\\\underline{0.8687}\\$\mathbf{0.8735}$}
&\makecell[c]{3.9004\\3.9021\\1.3848\\1.1007\\\underline{1.0495}\\$\mathbf{1.0254}$}
&\makecell[c]{6.9659\\6.7890\\2.8742\\2.9632\\\underline{2.5738}\\$\mathbf{2.4027}$}\\
				\hline
			\end{tabular}
		\end{lrbox}
		\centering
		\caption{PSNR, SSIM,  MSE and CIEde2000 values of  the restored color images in Section \ref{exam1}.}
		\scalebox{0.85}{\usebox{\tablebox}}
		\label{tab:modelcom_sym}
	\end{table*}

\subsection{Symmetric  cross-channel deblurring problem}\label{exam1}
	In this example, we consider the case where all sub-blur kernels are identical and the weight matrix is symmetric. The degraded images used for testing are generated by applying the  cross-channel blur $\Kq=K\odot W$  with  Gaussian blur  $K=ones(3,3)\otimes (G,5,5)$ and  weight matrix 
	$$
	W=\left [
	\begin{array}{ccc}
		0.7&0.15& 0.15\\
		0.15&0.7&0.15\\
		0.15&0.15&0.7
	\end{array}
	\right]
	$$
and  Gaussian noise with standard deviation $\sigma=0.01$  to the clean color images.  The observed images after degradation process are listed in the first column of 
 Figure $\ref{fig:modelcom_sym}$.

	We apply six compared method to the blurred and noised color images and use the newly proposed  L-surface method to  choose optimal model parameters.
The evaluation criteria  values of the restored color images are listed in Table \ref{tab:modelcom_sym} with  the best values in bold and  the secondly best ones underlined. 
One can see that the newly proposed CSTV achieves at  the best PSNR, SSIM and MSE  values among six methods.  
SVTV performs slightly worse than CSTV, and in particular, it outperforms CSTV on the CIEde2000 values of two testing color images 'statues'  and  'penguin'.

The restorations of six methods are given in Figure $\ref{fig:modelcom_sym}$.   These numerical results indicate that all of them successfully solved the symmetric cross-channel deblurring problem and recovered color images at a high level.   When looking at  details of colors and textures, one can see that CSTV performs better than other five methods. 
 The  color images restored by $\CTV_1$ and $\CTV_2$  
have distorted colors.  
The restorations of $\rm{\CTV_{split}}$ still have some artifacts;  for instance,  see the grass skirt detail in 'aborigine' and the watercress in 'bear'. 
$\MTV$ has achieved relatively good results both visually and numerically.
However,  $\MTV$ seems oversmoothing in some small details; see the flower texture in the upper right corner of the 'butterfly'. 
Under the same stopping criterion,  $\CSTV$  restores color images  with faithful colors and textures and  without artifacts and color artifacts. The best performance of $\CSTV$ numerically  verifies  its advantages in treating color pixel as a quaternion number and using quaternion operator splitting method.

To further highlight the differences of $\CSTV$ to  $\MTV$ and $\SVTV$, we enlarge the local features of their restorations in Figure \ref{zoom_sym}. The corresponding zoomed parts are indicated by red boxes in Figure \ref{fig:modelcom_sym}. The amplification of local features reveals that $\CSTV$ retains more texture detail; see the tip of a tree branch in 'waterfall', the texture of petals in 'butterfly', and the details of the letters in 'garden'. Additionally, we can see color artifacts at the edges of color images restored by $\MTV$, including the hair part in 'girl',  the left edge of the flower in 'butterfly', and the lower half of the fence in 'aborigine'.

\subsection{Asymmetric cross-channel deblurring problem}\label{exam2}
	In this example, we consider the restorations of  six compared methods when the sub-blur kernels are different and the weight matrix is asymmetric.
	The degraded images are generated by introducing the cross-channel blur $\Kq=K\odot W$ with
	\begin{equation*}
		K=\left [
		\begin{array}{ccc}
			(A, 5)&(M, 11, 45)&(M, 21, 90)\\
			(M, 11, 45),&(A, 5)&(G, 5, 5)\\
			(M, 21, 90)&(G, 5,5)&(A, 5)
		\end{array}
		\right ]
	\end{equation*}
and 
$$
	W=\left [
	\begin{array}{ccc}
		0.7&0.15& 0.15\\
		0.1&0.8&0.1\\
		0.05&0.05&0.9
	\end{array}
	\right ]
	$$
and  Gaussian noise with standard deviation $\sigma=0.01$ to original color images.  This blur operator is a mixture of  Gaussian blur $(G,5,5)$, motion blur $ (M,11,45)$ and average blur $(A,5)$.

	\begin{table*}[htbp]
		\begin{lrbox}{\tablebox}
			\begin{tabular}{|l|c|c|c|c|c||l|c|c|c|c|c|}
				\hline
				\makecell[c]{Observed image\\(PSNR~SSIM)}
				&Method  &{\tt PSNR} & {\tt SSIM} & \makecell[c]{{\tt MSE} \\(\num{1e-3})} & {\tt CIEde2000}&\makecell[c]{Observed image\\(PSNR~SSIM)}
				&Method  &{\tt PSNR} & {\tt SSIM} & \makecell[c]{{\tt MSE} \\(\num{1e-3})} & {\tt CIEde2000}\\  \hline
				
\makecell[c]{statues\\(22.8211~0.4890)}
&\makecell[c]{$\CTV_1$\\$\CTV_2$\\$\rm{\CTV_{split}}$\\$\MTV$\\$\SVTV$\\ $\CSTV$}
&\makecell[c]{24.8674\\24.8129\\18.6735\\26.7807\\\underline{27.4255}\\$\mathbf{27.4570}$}
&\makecell[c]{0.6587\\0.6510\\0.6336\\0.7275\\\underline{0.7674}\\$\mathbf{0.7703}$}
&\makecell[c]{3.2603\\3.3015\\13.572\\2.0986\\\underline{1.8091}\\$\mathbf{1.7960}$}
&\makecell[c]{5.7584\\5.4356\\14.4771\\4.3106\\$\mathbf{3.0850}$\\\underline{3.1961}}&
				
\makecell[c]{waterfall\\(21.87546~0.5683)}
&\makecell[c]{$\CTV_1$\\$\CTV_2$\\$\rm{\CTV_{split}}$\\$\MTV$\\$\SVTV$\\ $\CSTV$}
&\makecell[c]{23.5651\\23.5176\\16.2687\\26.3129\\\underline{26.9149}\\$\mathbf{27.0396}$}
&\makecell[c]{0.6927\\0.7005\\0.6819\\0.7962\\$\mathbf{0.8197}$\\\underline{0.8152}}
&\makecell[c]{4.4004\\4.4488\\23.612\\2.3373\\\underline{2.0347}\\$\mathbf{1.9772}$}
&\makecell[c]{6.0229\\5.8441\\15.3867\\3.8289\\\underline{3.1820}\\$\mathbf{3.1800}$}\\
				\hline
				
\makecell[c]{aborigine\\(22.3385~0.5643)}
&\makecell[c]{$\CTV_1$\\$\CTV_2$\\$\rm{\CTV_{split}}$\\$\MTV$\\$\SVTV$\\ $\CSTV$}
&\makecell[c]{23.8554\\23.8243\\15.1137\\27.7450\\\underline{28.3032}\\$\mathbf{28.4469}$}
&\makecell[c]{0.7054\\0.7091\\0.6704\\0.8013\\\underline{0.8261}\\$\mathbf{0.8268}$}
&\makecell[c]{4.1159\\4.1454\\30.805\\1.6808\\\underline{1.4780}\\$\mathbf{1.4300}$}
&\makecell[c]{6.0587\\5.8796\\15.6282\\3.4535\\\underline{2.7616}\\$\mathbf{2.7396}$}&
				
\makecell[c]{bear\\(20.4969~0.5420)}
&\makecell[c]{$\CTV_1$\\$\CTV_2$\\$\rm{\CTV_{split}}$\\$\MTV$\\$\SVTV$\\ $\CSTV$}
&\makecell[c]{22.1822\\22.1152\\14.9329\\24.5403\\\underline{25.1865}\\$\mathbf{25.3409}$}
&\makecell[c]{0.6965\\0.6947\\0.6858\\0.7876\\\underline{0.8173}\\$\mathbf{0.8188}$}
&\makecell[c]{6.0504\\6.1443\\32.115\\3.5153\\\underline{3.0293}\\$\mathbf{2.9234}$}
&\makecell[c]{6.6658\\6.5229\\18.1410\\4.5132\\\underline{3.7226}\\$\mathbf{3.5917}$}\\
				\hline
				
\makecell[c]{penguin\\(21.5152~0.5904)}
&\makecell[c]{$\CTV_1$\\$\CTV_2$\\$\rm{\CTV_{split}}$\\$\MTV$\\$\SVTV$\\ $\CSTV$}
&\makecell[c]{23.6758\\23.6289\\16.1198\\26.3597\\\underline{26.8866}\\$\mathbf{26.9257}$}
&\makecell[c]{0.7131\\0.7171\\0.7181\\0.7989\\\underline{0.8171}\\$\mathbf{0.8172}$}
&\makecell[c]{4.2896\\4.3362\\24.435\\2.3122\\\underline{2.0480}\\$\mathbf{2.0297}$}
&\makecell[c]{6.1883\\5.7965\\20.1197\\4.9455\\$\mathbf{3.4193}$\\\underline{3.5165}}&
				
\makecell[c]{butterfly\\(21.5967~0.6596)}
&\makecell[c]{$\CTV_1$\\$\CTV_2$\\$\rm{\CTV_{split}}$\\$\MTV$\\$\SVTV$\\ $\CSTV$}
&\makecell[c]{23.6836\\23.6712\\15.4844\\28.2452\\\underline{28.4956}\\$\mathbf{28.5637}$}
&\makecell[c]{0.7609\\0.7694\\0.7548\\0.8625\\\underline{0.8673}\\$\mathbf{0.8730}$}
&\makecell[c]{4.2819\\4.2942\\2.8285\\1.4979\\\underline{1.4140}\\$\mathbf{1.3920}$}
&\makecell[c]{6.1226\\5.9530\\14.9422\\3.1680\\\underline{2.7874}\\$\mathbf{2.5219}$}\\
				\hline
\makecell[c]{garden\\(19.4949~0.5044)}
&\makecell[c]{$\CTV_1$\\$\CTV_2$\\$\rm{\CTV_{split}}$\\$\MTV$\\$\SVTV$\\ $\CSTV$}
&\makecell[c]{21.5268\\21.4425\\16.6254\\25.0593\\\underline{25.7302}\\$\mathbf{25.9486}$}
&\makecell[c]{0.6970\\0.6941\\0.6942\\0.8270\\$\mathbf{0.8510}$\\\underline{0.8459}}
&\makecell[c]{7.0359\\7.1738\\21.750\\3.1194\\\underline{2.6729}\\$\mathbf{2.5418}$}
&\makecell[c]{7.5611\\7.3168\\18.7589\\5.2695\\\underline{4.4351}\\$\mathbf{4.1440}$}&
				
\makecell[c]{girl\\(23.0120~0.6945)}
&\makecell[c]{$\CTV_1$\\$\CTV_2$\\$\rm{\CTV_{split}}$\\$\MTV$\\$\SVTV$\\ $\CSTV$}
&\makecell[c]{24.6124\\24.6185\\16.4629\\29.3705\\\underline{29.8001}\\$\mathbf{29.8680}$}
&\makecell[c]{0.7778\\0.7885\\0.7647\\0.8552\\\underline{0.8665}\\$\mathbf{0.8695}$}
&\makecell[c]{3.4503\\3.4526\\22.579\\1.1560\\\underline{1.0471}\\$\mathbf{1.0309}$}
&\makecell[c]{6.2335\\6.0251\\18.9103\\3.1128\\\underline{2.5057}\\$\mathbf{2.4164}$}\\
				\hline
			\end{tabular}
		\end{lrbox}
		\centering
		\caption{PSNR, SSIM,  MSE and CIEde2000 values of  the restored color image in Section \ref{exam2}.}
		\scalebox{0.85}{\usebox{\tablebox}}
		\label{tab:modelcom}
	\end{table*}

\begin{figure}[!t]
		\centering
		\includegraphics[width=0.48\textwidth]{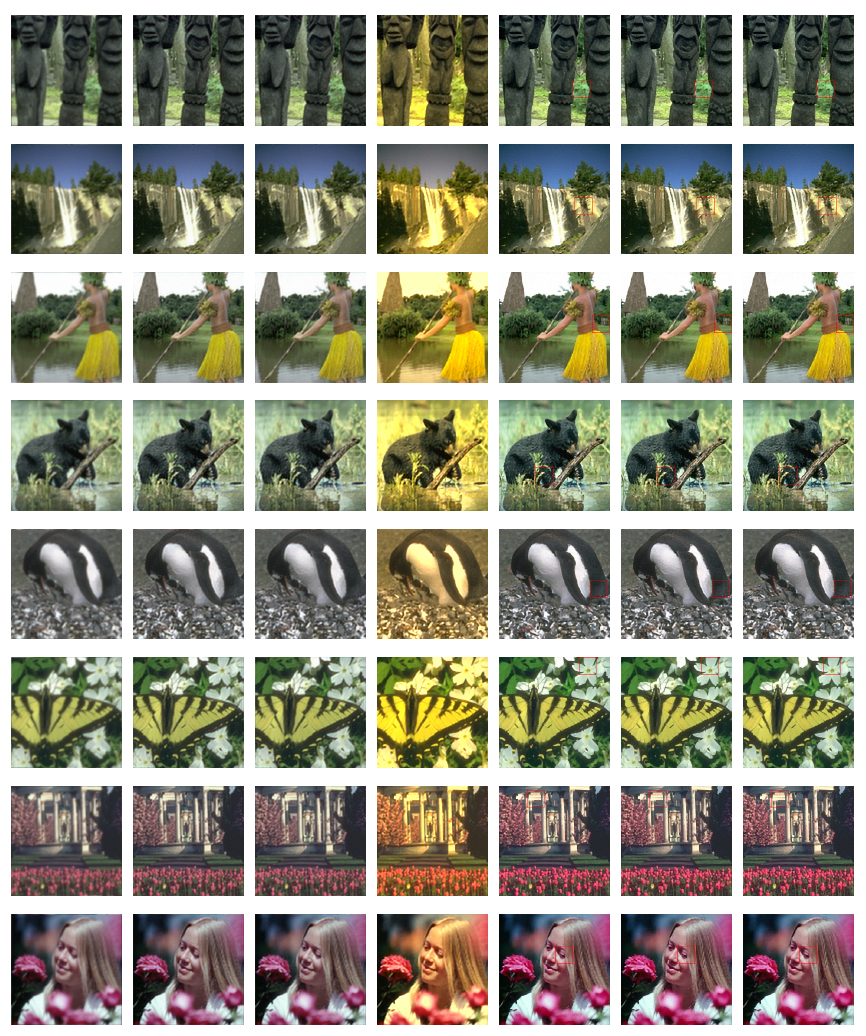}
		\caption{Visual comparison of six methods in asymmetric cross-channel deblurring. From left to right: Observed images and restorations by  $\CTV_1$, $\CTV_2$,  
$\rm{\CTV_{split}}$, $\MTV$, $\SVTV$, CSTV. }
		\label{fig:modelcom}
	\end{figure}
	
	\begin{figure}[!t]
		\centering
		\includegraphics[width=0.48\textwidth]{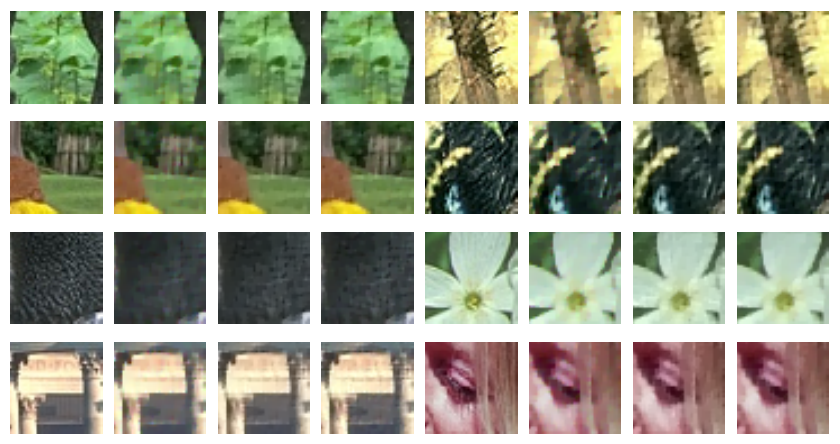}
		\caption{Corresponding zooming parts of the original color images and the restorations of $\MTV$, $\SVTV$ and $\CSTV$   given in Figure \ref{fig:modelcom}. 
		}
		\label{zoom}
	\end{figure}

In Figure \ref{fig:modelcom}, we display the observed color images and the restored color images by the six methods.  These numerical results indicate that the asymmetric cross-channel deblurring problem is difficult for $\CTV_1$,  $\CTV_2$  and $\rm{\CTV_{split}}$.   Visually,  the restorations of  $\CTV_1$ and  $\CTV_2$ still contain  cross-channel blurred color pixels and a slight color shift.  For example, the texture of the grass skirt in 'aborigine' is relatively smooth and darker overall.  The color images restored  by $\rm{\CTV_{split}}$ have a significant color shift in this design of asymmetric weight matrix and different sub-blur kernels. 
Fortunately, the other three methods  $\MTV$, $\SVTV$ and $\CSTV$ are able to solve such asymmetric cross-channel deblurring problem.  We zoom in the local details of the restorations of them (the zoomed areas are indicated by red boxes in Figure \ref{fig:modelcom}) in Figure \ref{zoom}. Visually,  the  proposed $\CSTV$  achieves the best restorations that preserve the geometric textures and color information. 
 From the comparison of the skin detail in 'aborigine' and the petal textures  in 'butterfly', we can see that $\CSTV$ produces better texture features, while the $\MTV$  tend to oversmooth. Also, $\CSTV$ produces superior results in terms of color fidelity judging from the branch edges in 'waterfall' and the petal edges in 'butterfly', where the $\MTV$ tends to have color artifacts in the sharply changing areas of the images.

We present the PSNR, SSIM, MSE, and CIEde2000 values of the color images restored by six compared methods  in Table \ref{tab:modelcom} with  the best values in bold and  the secondly best ones underlined. One can evidently observe that the new approach $\CSTV$ achieves the best average values of PSNR, SSIM,  MSE and CIEde2000.  Especially,  $\CSTV$ has an average minimum reduction of  CIEde2000 around $0.64$ compared with the other five methods.
From the comparison between Tables \ref{tab:modelcom_sym} and \ref{tab:modelcom}, it is worth emphasizing that the restoration numerical results of  $\MTV$  are worse than the corresponding results in Example \ref{exam1} in terms of indicators such as CIEde2000, while those of $\CSTV$ are more stable for the restorations of different blur kernels. This means that $\MTV$ is more sensitive to the design of cross-channel blur kernel and  $\CSTV$ is more general and stable. These numerical results support the efficiency, stableness and superiority of the newly proposed $\CSTV$.

 In addition, we compare the performance of above methods at different levels of noise.  We take `butterfly' as the testing color image and  set the noise levels as  $0.05$, $0.1$ and $0.5$. 
	The numerical results are given in Table \ref{tab:noiselevel}. Note that the notation `-`  means that  $\CTV_1$, $\CTV_2$ and $\MTV$ fail to compute  acceptable restorations  in the third case. 
	One can see that the newly proposed $\CSTV$  performs better than the other compared methods  at color image deblurring with different levels of noise.  The restored color images and parameter values are presented in the supplementary material.

		\begin{table}[htbp]
		\begin{lrbox}{\tablebox}
			\begin{tabular}{|l|c|c|c|c|c|}
				\hline
				Noise level&Method
				&{\tt PSNR} & {\tt SSIM} & \makecell[c]{{\tt MSE} \\(\num{1e-3})} & {\tt CIEde2000}\\  \hline
\makecell[c]{$0.05$}
				&\makecell[c]{$\CTV_1$\\$\CTV_2$\\$\MTV$\\$\SVTV$\\ $\CSTV$}
				&\makecell[c]{20.8483\\21.1753\\24.7291\\\underline{24.8010}\\\bf{24.9349}}
				&\makecell[c]{0.4866\\0.5092\\0.7493\\\underline{0.7627}\\\bf{0.7631}}
				&\makecell[c]{8.2260\\7.6290\\3.3660\\\underline{3.3110}\\\bf{3.2100}}
				&\makecell[c]{12.3094\\11.6168\\4.9102\\\underline{4.2139}\\\bf{4.1395}}
				\\  \hline
\makecell[c]{$0.1$}
				&\makecell[c]{$\CTV_1$\\$\CTV_2$\\$\MTV$\\$\SVTV$\\ $\CSTV$}
				&\makecell[c]{16.9792\\17.3635\\20.5315\\\underline{23.2933}\\\bf{23.4325}}
				&\makecell[c]{0.2985\\0.3126\\0.4967\\\underline{0.6974}\\\bf{0.7009}}
				&\makecell[c]{20.0480\\18.3500\\8.8480\\\underline{4.4850}\\\bf{4.5370}}
				&\makecell[c]{19.3420\\18.6192\\11.2231\\\bf{5.2823}\\\underline{5.3292}}
				\\  \hline
\makecell[c]{$0.5$}
				&\makecell[c]{$\CTV_1$\\$\CTV_2$\\$\MTV$\\$\SVTV$\\ $\CSTV$}
				&\makecell[c]{-\\-\\-\\\underline{18.3643}\\\bf{18.5622}}
				&\makecell[c]{-\\-\\-\\\underline{0.4124}\\\bf{0.4221}}
				&\makecell[c]{-\\-\\-\\\underline{14.5740}\\\bf{13.9240}}
				&\makecell[c]{-\\-\\-\\\underline{13.4633}\\\bf{13.3424}}
				\\  \hline
			\end{tabular}
		\end{lrbox}
		\centering
		\caption{Numerical values for the cases with different levels of noise in Section \ref{exam2}. 
		}
		\scalebox{0.85}{\usebox{\tablebox}}
		\label{tab:noiselevel}
\end{table}

\subsection{Analysis of cross-channel deblurring ability}\label{exam3}
In this example, we elucidate the benefits of using quaternion characterization of blurring and  quaternion operator splitting method. 
The observed image ('butterfly')  is solely  affected by a blurring operator, without any distortion caused by noise. The chosen blur operators are  Gaussian blur $ones(3,3)\otimes (G,5,5)$, motion blur $ones(3,3)\otimes (M,11,45)$ and average blur $ones(3,3)\otimes (A,5)$ with weight matrix
$$
W=\left [
\begin{array}{ccc}
	0.7&0.15& 0.15\\
	0.15&0.7&0.15\\
	0.15&0.15&0.7
\end{array}
\right ].
$$

Our discussion encompasses two facets:  regularization and  algorithm. We denote the CSTV regularization model \eqref{m:CSmodel} with quaternion operator splitting solver (Algorithm \ref{code:Qcstv}) by $\CSTV_\QGMRES$.  
From the  regularization aspect,  we test two special cases of the CSTV regularization model \eqref{m:CSmodel}: $\SVTV_\QGMRES$  $(\lambda_1>0, \lambda_2=0)$  and   $\CTV_\QGMRES$ $(\lambda_1=0, \lambda_2>0)$.  
For the algorithm aspect, we change the fidelity term  \eqref{e:fidelity} to the fidelity term \eqref{e:fidelity0} and solve  model \eqref{m:CSmodel} by the real ADMM method with GMRES solver.  This version uses only real operator algorithms and  does not use quaternion operator algorithms.  
We denote it   by $\CSTV_\GMRES$ and  denote its two special cases by $\SVTV_\GMRES$  $(\lambda_1>0, \lambda_2=0)$  and   $\CTV_\GMRES$ $(\lambda_1=0, \lambda_2>0)$.

The numerical results in the aforementioned six cases are presented in Table \ref{tab:deblur}.
One can easily find that the performance of the proposed $\CSTV$ regularization model is significantly higher than the corresponding results of the single ($\SVTV$ or $\CTV$) regularization model in each evaluation index.   For $\CTV_\GMRES$, $\SVTV_\GMRES$ and $\CSTV_\GMRES$, the PSNR value can be increased by approximately 1.4, the SSIM value can be improved by up to 7.27$\%$, the MSE value can be reduced by as much as 31.5$\%$, and the CIEde2000 value can be decreased by up to 16.9$\%$. For  $\CTV_\QGMRES$, $\SVTV_\QGMRES$ and $\CSTV_\QGMRES$,  the PSNR value can be increased by an average of approximately 1, the SSIM value can be improved by up to 3.57$\%$, the MSE value can be reduced by as much as 27.5$\%$, and the CIEde2000 value can be decreased by up to 10.9$\%$. 

When examining the algorithm for model resolution, it becomes visually evident from Table \ref{tab:deblur} that the quaternion operation process ($\CTV_\QGMRES$, $\SVTV_\QGMRES$ and $\CSTV_\QGMRES$) significantly enhances the corresponding evaluation metrics of color images deblurred by the real operation process ($\CTV_\GMRES$, $\SVTV_\GMRES$ and $\CSTV_\GMRES$).  Averagely, the PSNR value sees an increase of 2.4, the SSIM value improves by 0.06, the MSE value decreases by 1.5e-4, and the CIEde2000 value reduces by 0.72.

\begin{figure}[htbp]
	\centering
	\includegraphics[width=0.48\textwidth]{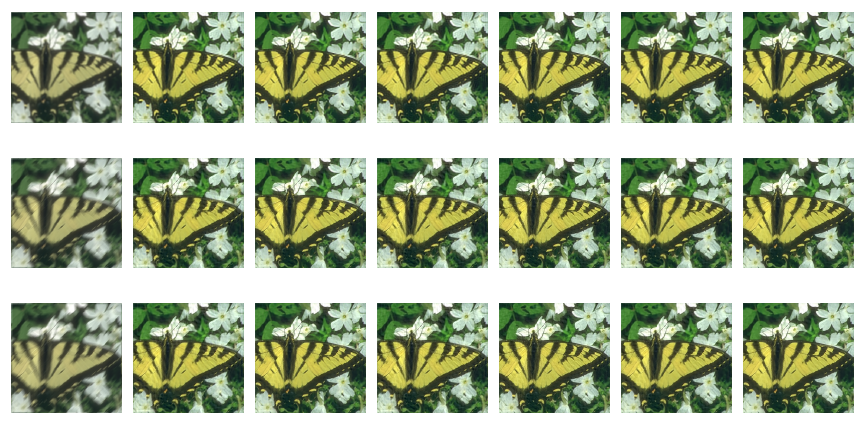}
	\caption{From left to right: the observed color images  and the deblurred color images by  $\CTV_{\GMRES}$, $\SVTV_{\GMRES}$, $\CSTV_{\GMRES}$,  $\CTV_{\QGMRES}$, $\SVTV_{\QGMRES}$ and $\CSTV_{\QGMRES}$.  The first row is for Gaussian blur, the second row is for motion blur, and the third row is average blur.
}
	\label{f:deblur}
\end{figure}

\begin{figure}[!t]
	\centering
\includegraphics[width=0.48\textwidth]{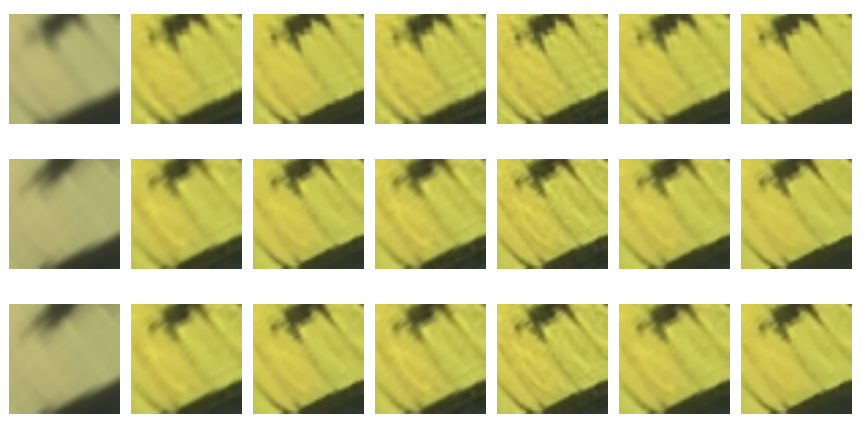}	
\caption{Zoomed parts of color images in Figure \ref{f:deblur}.
}
\label{f:deblur_zoom}
\end{figure}

The deblurred color images in the aforementioned six cases are presented  in Figure \ref{f:deblur} and the zoomed parts are shown in Figure \ref{f:deblur_zoom}.
One can find that  color images  deblurred by  $\CSTV_\QGMRES$ can better restore the details, such as maintaining the continuity of the lines. Moreover, the feature details restored by $\CSTV_\QGMRES$ under quaternion operations  become even more striking than those by $\CSTV_\GMRES$ under real operations.

\begin{figure}[!t]
	
	\centering
	
	\includegraphics[width=0.40\textwidth]{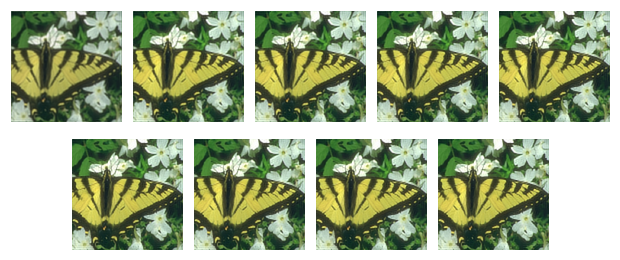}
	\caption{Restorations at iteration step $k$  (k=1, 4, 7, 10, 13, 16, 19, 22, 24) of the deblurring process by ${\CSTV_\GMRES}$.
	}
	\label{f:sctv_gmres_step}
\end{figure}

\begin{figure}[!t]
	
	\centering
	\includegraphics[width=0.48\textwidth]{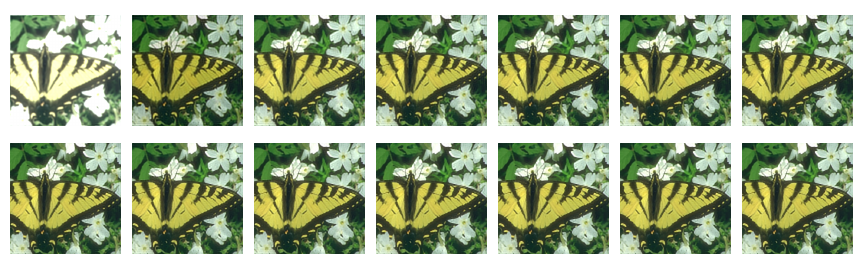}
	\caption{Restorations at iteration step $k$  (k=1, 4, 7, 10, 13, 16, 19, 22, 25, 28, 31, 34, 37, 41) of the deblurring process by ${\CSTV_\QGMRES}$.
	}
	\label{f:sctv_qgmres_step}
\end{figure}

\begin{figure}[!t]
	\subfigure{\includegraphics[width=0.24\textwidth,height=0.16\textwidth]{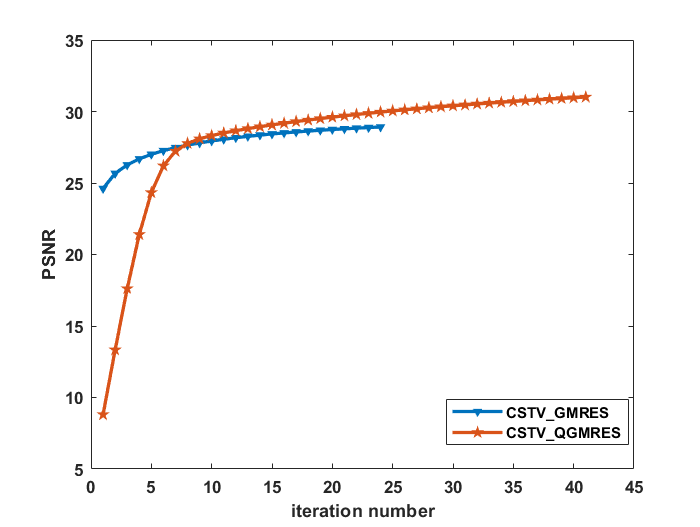}}
	\subfigure{\includegraphics[width=0.24\textwidth,height=0.16\textwidth]{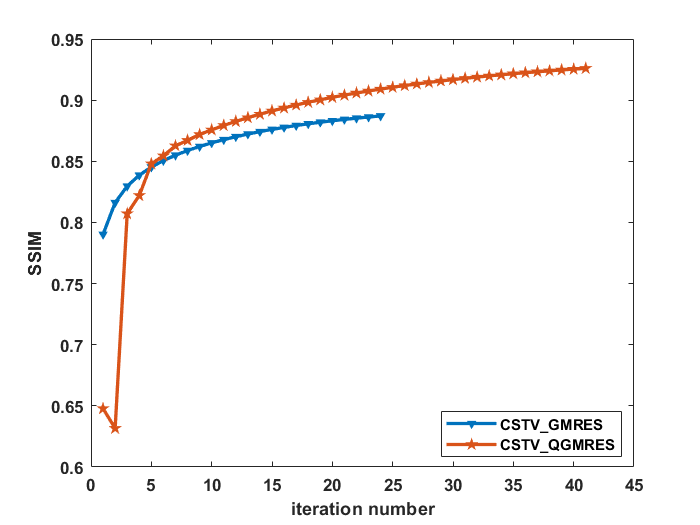}}
	\subfigure{\includegraphics[width=0.24\textwidth,height=0.16\textwidth]{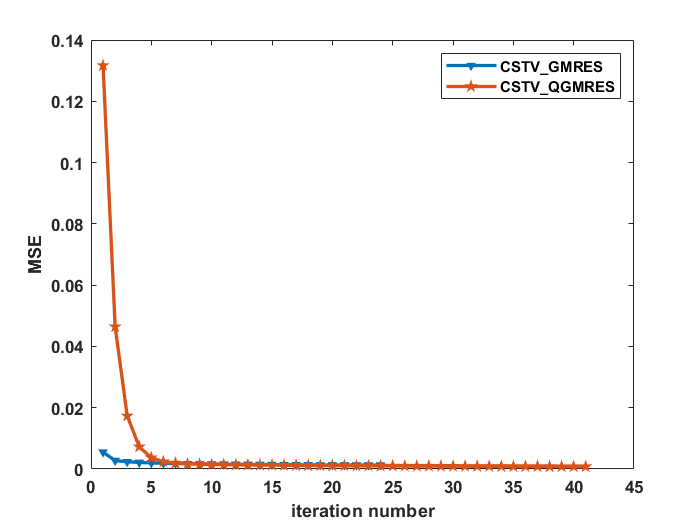}}
	\subfigure{\includegraphics[width=0.24\textwidth,height=0.16\textwidth]{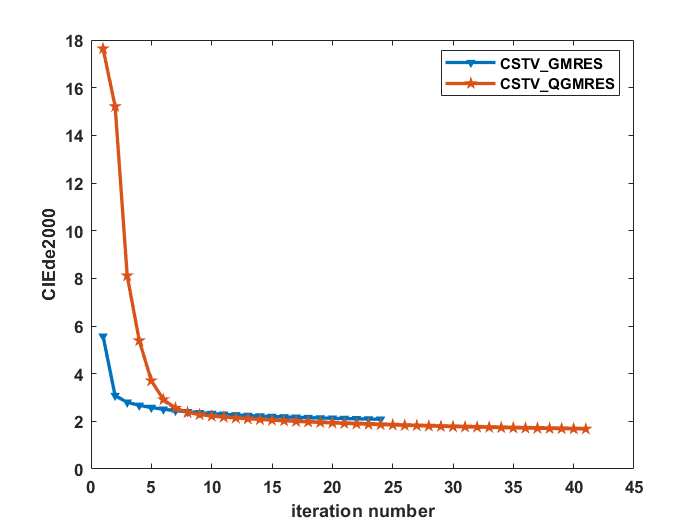}}
	
\caption{PSNR, SSIM, MSE and CIEde2000 values of color images computed by ${\CSTV_\GMRES}$ and ${\CSTV_\QGMRES}$ at iteration step $k$.
}
\label{f:sctv_step}
\end{figure}

In order to further illustrate the advantages of  algorithms,  we show the step-by-step results $\uq_k$ of the restoration process 
  using $\CSTV_{\GMRES}$ and $\CSTV_{\QGMRES}$ in Figures \ref{f:sctv_gmres_step} and \ref{f:sctv_qgmres_step}.   
 In Figure \ref{f:sctv_step}, we show the comparison results of PSNR, SSIM, MSE and CIEde2000 values corresponding to the step-by-step result $\uq_k$ during the restoration process.  $\CSTV_{\QGMRES}$ converges in $41$ iteration steps and $\CSTV_{\GMRES}$ convergences in $24$ iteration steps.   The average CPU time for a single-step iteration of $\CSTV_{\QGMRES}$  is 2.7931 seconds and that of $\CSTV_{\GMRES}$ is 1.4757 seconds.
 However,  the restoration of $\CSTV_{\QGMRES}$ 
achieve higher quality  than that of $\CSTV_{\GMRES}$. At the starting, the PSNR, SSIM, MSE and CIEde2000 values of the former are not better than those of the latter.  After a few of iterations, $\CSTV_{\QGMRES}$  performs much better than $\CSTV_{\GMRES}$.

For the above analysis, we  can make two key assertions: 
\begin{itemize}
\item Firstly, the deblurring efficiency of the CSTV regularization model is markedly superior to that of the CTV or SVTV regularization model under identical parameter settings. 
\item Secondly,  the CTV,  SVTV and CSTV regularization models  with applying  quaternion blur operation $\Qq$ significantly outperform their counterparts with applying  real blur operation $\Kq$ under the same parameter settings. 
\end{itemize}

		\begin{table}[htbp]
		\begin{lrbox}{\tablebox}
			\begin{tabular}{|l|c|c|c|c|c|}
				\hline
				Blur type&Method  &{\tt PSNR} & {\tt SSIM} & \makecell[c]{{\tt MSE} \\(\num{1e-3})} & {\tt CIEde2000}\\  \hline
				\makecell[c]{Gaussian}
				&\makecell[c]{$\CTV_{\GMRES}$\\$\SVTV_{\GMRES}$\\$\CSTV_{\GMRES}$}
				&\makecell[c]{27.5742\\27.9638\\$\mathbf{28.9120}$}
				&\makecell[c]{0.8566\\0.8598\\$\mathbf{0.8878}$}
				&\makecell[c]{1.7482\\1.5981\\$\mathbf{1.2847}$}
				&\makecell[c]{2.4202\\2.3634\\$\mathbf{2.0897}$}
				\\  \hline
				\makecell[c]{Gaussian}
				&\makecell[c]{$\CTV_{\QGMRES}$\\$\SVTV_{\QGMRES}$\\$\CSTV_{\QGMRES}$}
				&\makecell[c]{30.1270\\30.2314\\$\mathbf{30.9028}$}
				&\makecell[c]{0.9109\\0.9040\\$\mathbf{0.9230}$}
				&\makecell[c]{0.97119\\0.94812\\$\mathbf{0.81231}$}
				&\makecell[c]{1.8073\\1.8845\\$\mathbf{1.7156}$}
				\\  \hline
				
				\makecell[c]{Motion}
				&\makecell[c]{$\CTV_{\GMRES}$\\$\SVTV_{\GMRES}$\\$\CSTV_{\GMRES}$}
				&\makecell[c]{25.8199\\26.4653\\$\mathbf{27.5577}$}
				&\makecell[c]{0.8111\\0.8263\\$\mathbf{0.8570}$}
				&\makecell[c]{2.6183\\2.2567\\$\mathbf{1.7948}$}
				&\makecell[c]{3.2780\\3.0875\\$\mathbf{2.7253}$}
				\\  \hline
				\makecell[c]{Motion}
				&\makecell[c]{$\CTV_{\QGMRES}$\\$\SVTV_{\QGMRES}$\\$\CSTV_{\QGMRES}$}
				&\makecell[c]{29.0774\\29.6386\\$\mathbf{30.4742}$}
				&\makecell[c]{0.8897\\0.8914\\$\mathbf{0.9129}$}
				&\makecell[c]{1.2367\\1.0868\\$\mathbf{0.89655}$}
				&\makecell[c]{2.2995\\2.2388\\$\mathbf{2.0485}$}
				\\  \hline
				
				\makecell[c]{Average}
				&\makecell[c]{$\CTV_{\GMRES}$\\$\SVTV_{\GMRES}$\\$\CSTV_{\GMRES}$}
				&\makecell[c]{23.5221\\23.9698\\$\mathbf{24.6607}$}
				&\makecell[c]{0.6974\\0.7142\\$\mathbf{0.7481}$}
				&\makecell[c]{4.4442\\4.0088\\$\mathbf{3.4193}$}
				&\makecell[c]{4.2445\\4.0452\\$\mathbf{3.6768}$}
				\\  \hline
				\makecell[c]{Average}
				&\makecell[c]{$\CTV_{\QGMRES}$\\$\SVTV_{\QGMRES}$\\$\CSTV_{\QGMRES}$}
				&\makecell[c]{25.5468\\25.8810\\$\mathbf{26.2855}$}
				&\makecell[c]{0.7853\\0.7869\\$\mathbf{0.8124}$}
				&\makecell[c]{2.7882\\2.5816\\$\mathbf{2.3521}$}
				&\makecell[c]{3.2664\\3.2145\\$\mathbf{3.0027}$}
				\\  \hline
			\end{tabular}
		\end{lrbox}
		\centering
		\caption{PSNR, SSIM,  MSE and CIEde2000 values of  the deblurred color images in Section \ref{exam3}.}
		\scalebox{0.85}{\usebox{\tablebox}}
		\label{tab:deblur}
\end{table}

\section{Conclusion}\label{s:conclusion}
In this paper, we have proposed a novel CSTV regularization model for cross-channel deblurring problem of color images. The proposed model contains a new joint  regularization term that exploits the complementary information of two or more different color spaces such as RGB and HSV.  It also has a new  fidelity term  that is characterized by quaternion  blur operators.   A new fast and  stable quaternion operator splitting algorithm is  proposed to solve the  proposed CSTV regularization model.  The advantages of the newly defined  regularization and fidelity terms are explained in theory and are verified in numerical examples.  

The proposed model and algorithm form a general framework that provides new ideas to improve many color TV-based models for color image restoration.  We believe that there is still room for further optimization of our model by introducing the variational contrast enhancement method in \cite{pabst17}. 
In future,  we will consider this valuable topic and apply our method to study the blind cross-channel deblurring problem.

	\section*{Acknowledgment}
We are grateful to the editor and the anonymous referees  for their  wonderful comments and helpful suggestions. 
The first author would also like to thank Prof. Jin Cheng and Prof. Jijun Liu for their excellent comments on this work at the conference “Intelligent Computing for Inverse Problems and Data Science (July 29-August 2, 2024,  TSIMF in Sanya)”.

\newpage
\begin{IEEEbiography}{Zhigang Jia}
received the Ph.D. degree in mathematics from East China Normal University, Shanghai, China, in 2009.  
He is currently a Professor in Mathematics, School of Mathematics and Statistics, Jiangsu Normal University, Xuzhou, China. His current research interests include numerical linear algebra,  data mining, quaternion matrix computations, color image recognition, medical image processing and imaging science.
\end{IEEEbiography}

\begin{IEEEbiography}[]{Yuelian Xiang}  received the M.Sc. degree from Jiangsu Normal University, China, in 2024. She is currently working at Wuhan Vocational College of Software and Engineering, China. Her current research interests include quaternion matrix computation, color image processing, and machine learning.
\end{IEEEbiography}

\begin{IEEEbiography}[]{Meixiang Zhao}
received the Ph.D. degree in School of
Information and Control Engineering, with a major in
control theory and control engineering. Her current
research interests include computation intelligence in
many-objective optimization, numerical mathematics,
machine learning, and applications in face recognition,
video processing, big data processing and analysis.
\end{IEEEbiography}

\begin{IEEEbiography}[]{Tingting Wu} received the B.S. and Ph.D. degrees in mathematics from Hunan University, Changsha, China, in 2006 and 2011, respectively. From 2015 to 2018, she was a Postdoctoral Researcher with the
School of Mathematical Sciences, Nanjing Normal University, Nanjing, China.
 She is currently a Professor with the School of Science, Nanjing University of Posts and Telecommunications, Nanjing. Her research interests include variational methods for image processing and computer vision, optimization methods and their applications in sparse recovery, and regularized inverse problems.
\end{IEEEbiography}

\begin{IEEEbiography}[]{Michael K. Ng}
received the B.Sc. and M.Phil. degrees from The University of Hong Kong, Hong Kong, in 1990 and 1992,
respectively, and the Ph.D. degree from The Chinese University of Hong Kong, Hong Kong, in 1995.
From 1995 to 1997, he was a Research Fellow with the Computer Sciences Laboratory, The Australian National University, Canberra, ACT, Australia. 
He was an Assistant Professor/an Associate Professor with The University of Hong Kong from 1997 to 2005. 
From 2006 to 2019, he was a Professor/the Chair Professor of the Department of Mathematics, Hong Kong Baptist University, Hong Kong. 
From 2020 to 2023, he was the Chair Professor of the Research Division of Mathematical and Statistical Science, The University of Hong Kong. 
 He is a Professor/the Chair Professor of the Department of Mathematics, the Dean of Science, Hong Kong Baptist University, Hong Kong. 
 His research interests include bioinformatics, image processing, scientific computing, and data mining. Dr. Ng is selected for the 2017 class of fellows of the Society for Industrial and Applied Mathematics. He received the Feng Kang Prize for his significant contributions to scientific computing.
  He serves as the editorial board member for several international journals.
\end{IEEEbiography}

\newpage

\appendix


Due to limited pages, we illustrate the solvability of the model and the parameter selection, and  present the details of the subproblem  of solving  CSTV regularization model in this supplementary material.

\subsection{Additional explanation of Figure 3 in the main body}

For more clarity, we explain the operational details of Figure 3 in the main body. For the mesh subdivision case, we set the mesh search step to $5\times 10^{-5}$. This searching step is sufficient to explore the superiority of the coupling of the regularization terms without adding additional tuning time. The weighted averaging process for the PSNR and SSIM surfaces is as follows. Firstly, the value ranges of the PSNR and SSIM surfaces are each normalized to the $ [0,1] $ interval. Secondly, equal weight factors of 0.5 are assigned to each of the normalized surfaces to achieve the weighted average.

\begin{figure}[htbp]
	\centering
	\subfigure[PSNR values in $\lambda_1=0$.]{\includegraphics[width=0.24\textwidth]{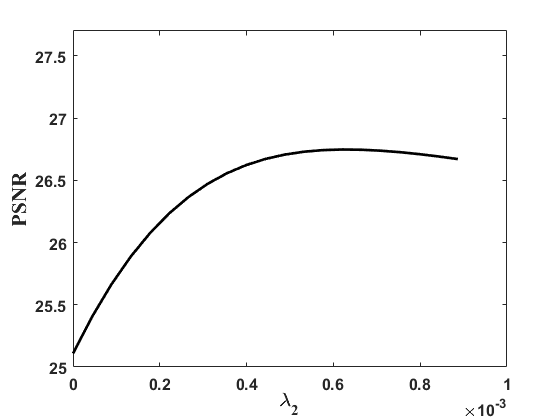}}
	\subfigure[SSIM values in $\lambda_1=0$.]{\includegraphics[width=0.24\textwidth]{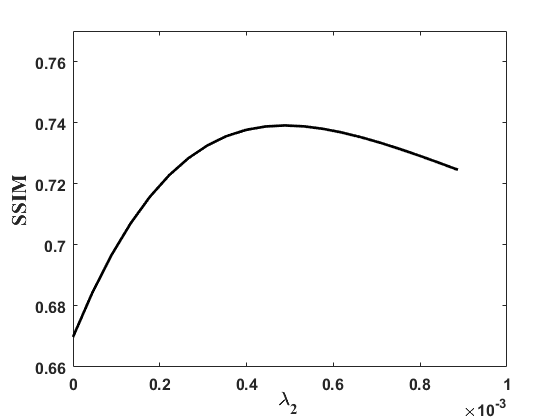}}\\
	\subfigure[PSNR values in $\lambda_2=0$.]{\includegraphics[width=0.24\textwidth]{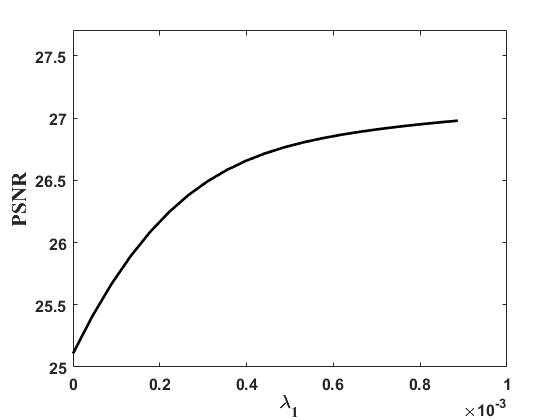}}
	\subfigure[SSIM values in $\lambda_2=0$.]{\includegraphics[width=0.24\textwidth]{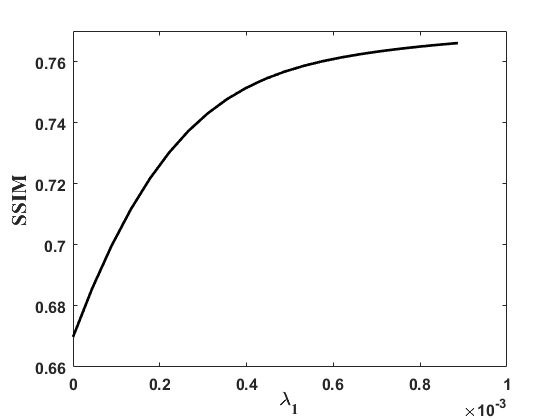}}
	\caption{The values of PSNR and SSIM for the restored image ‘statues’ in $\lambda_1=0$ or  $\lambda_2=0$.
	}
	\label{fig:lambda=0a}
\end{figure}

To demonstrate the superiority,  we plot  the PSNR and SSIM values  in Figure \ref{fig:lambda=0a} at boundary  $\lambda_1 = 0$ or $\lambda_2 = 0$.  We can see that  the PSNR and SSIM values  in Figure \ref{fig:lambda=0a} are significantly smaller than  $27.1903$ and $0.7704$ , respectively. 

\section{Proof of Theorem 1}
\begin{proof}
The color image cross-channel blurring process is  mathematically described as
\begin{align}\label{e:blurmodel0a}
	\hat{u}(x,y)& = \hat{K} \star u(x,y),
\end{align}
where
\begin{equation}\label{e:KWa}
\begin{aligned}
&\qquad \qquad\qquad\qquad \hat{K} =W\odot K=[K_{ij}\omega_{ij}]_{3\times 3},\\
	&K\!\!=\!\!\left[\!
	\begin{array}{ccc}
		K_{11}&K_{12}& K_{13}\\
		K_{21}&K_{22}&K_{23}\\
		K_{31}&K_{32}&K_{33}
	\end{array}\!
	\right ]\!\!, ~W\!\!=\!\!\left [\!
	\begin{array}{ccc}
		w_{11}&w_{12}& w_{13}\\
		w_{21}&w_{22}&w_{23}\\
		w_{31}&w_{32}&w_{33}
	\end{array}\!
	\right ]\!\!.
\end{aligned}
\end{equation}
That is, $\hat{u}_i(x,y)=\sum_{j=1}^3(K_{ij}w_{ij})\star u_j(x,y)$ for $i=1,2,3$.

Let $u_0(x,y)$  be a zero function.
 Then the cross-channel blurring process  can be embedded into a higher dimentional space, 
\begin{equation}\label{e:expandUa}
	\begin{aligned}
	\left [\begin{array}{cccc}
		\hat{u}_0(x,y)\\
		\hat{u}_1(x,y)\\
		\hat{u}_2(x,y)\\
		\hat{u}_3(x,y)
	\end{array}\right ] &=
	\left [\begin{array}{c:ccc}
		B_{11} & B_{12} &B_{13} & B_{14}\\
		\hdashline
		B_{21} &   		&		 &		  \\
		B_{31} & 		&\Kq&			\\
		B_{41} &   		&		 &
	\end{array}\right ] \star
	\left [\begin{array}{cccc}
		u_0(x,y)\\
		u_1(x,y)\\
		u_2(x,y)\\
		u_3(x,y)
	\end{array}\right ],
	\end{aligned}
\end{equation}
where  $B_{ij}$'s are arbitrary  operators of the same size of $K_{ij}$'s and $\hat{u}_0(x,y)=B_{12}\star u_1(x,y)+B_{1,3}\star u_2(x,y)+B_{14}\star u_3(x,y)$.
Denote the extended blurring operator by  $B$, that is, 
\begin{equation}\label{e:Bmatrixa}B=\left [\begin{array}{c:ccc}
		B_{11} & B_{12} &B_{13} & B_{14}\\
		\hdashline
		B_{21} &   		&		 &		  \\
		B_{31} & 		&\Kq&			\\
		B_{41} &   		&		 &
	\end{array}\right ].
\end{equation}
It can be splitted into a sum of a JRS-symmetric operator $Q$ and a $4 \times 4$ real block operator $R$,  according to the following rules:
\begin{equation}\label{e:KQRa}
	B=Q+R,
\end{equation}
where
\begin{equation}\label{matrix_spa}
	\begin{aligned}
		Q=\frac{1}{4}(B+J_4B{J_4}^T+R_4B{R_4}^T+S_4B{S_4}^T),\\
		R=\frac{1}{4}(3B-J_4B{J_4}^T-R_4B{R_4}^T-S_4B{S_4}^T),\\
	\end{aligned}
\end{equation}  
and $J_4, R_4, S_4$ are unitary operators defined in \cite{jwzc18a}. 
So we have 
$$B\star \left [\begin{array}{cccc}
		u_0(x,y)\\
		u_1(x,y)\\
		u_2(x,y)\\
		u_3(x,y)
	\end{array}\right]=Q\star \left [\begin{array}{cccc}
		u_0(x,y)\\
		u_1(x,y)\\
		u_2(x,y)\\
		u_3(x,y)
	\end{array}\right ]+R\star \left [\begin{array}{cccc}
		u_0(x,y)\\
		u_1(x,y)\\
		u_2(x,y)\\
		u_3(x,y)
	\end{array}\right ].$$

 Since these unitary operators satisfy
	$J_4J_4=-I_{4n}$, $J_4R_4=S_4$ and  $J_4S_4=-R_4$, 
we can easily derive
	\begin{equation*}
		J_4QJ_4^{T}=Q,~~R_4QR_4^{T}=Q,~~S_4QS_4^{T}=Q.
	\end{equation*}
From \cite{jwzc18a},  there exists a quaternion matrix 
\begin{equation}\label{e:Qmatrixa}\Qq=Q_0+Q_1\iq+Q_2\jq+Q_3\kq\end{equation}
such that 
\begin{equation}\label{e:Qrealcounterparta}
Q=\Re(\Qq),
\end{equation}
and
\begin{equation}\label{e:Q0123a}
	\begin{aligned}
		Q_0&=\frac{1}{4}(B_{11}+w_{11}K_{11}+w_{22}K_{22}+w_{33}K_{33}),\\
		Q_1&=\frac{1}{4}(B_{21}-B_{12}+w_{32}K_{32}-w_{23}K_{23}),\\
		Q_2&=\frac{1}{4}(B_{31}-B_{13}+w_{13}K_{13}-w_{31}K_{31}),\\
		Q_3&=\frac{1}{4}(B_{41}-B_{14}+w_{21}K_{21}-w_{12}K_{12}).
	\end{aligned}
\end{equation}

Define $\uq(x,y)=u_0(x,y)+u_1(x,y)\iq+u_2(x,y)\jq+u_3(x,y)\kq$ with $u_0(x,y)\equiv 0$. Then 
$$\Re_c(\uq(x,y))=[u_0(x,y), u_1(x,y), u_2(x,y),u_3(x,y)]^T.$$
So we have
$$Q\star\Re_c(\uq(x,y))=\Re(\Qq)\star \Re_c(\uq(x,y))=\Re_c(\Qq \textcircled{$\star$} \uq(x,y)).$$
Define 
$$\left[\begin{array}{rrrr}
			r_0(x,y)\\
			r_1(x,y)\\
			r_2(x,y)\\
			r_3(x,y)
		\end{array}\right]=R\star
\left[\begin{array}{rrrr}
			u_0(x,y)\\
			u_1(x,y)\\
			u_2(x,y)\\
			u_3(x,y)
		\end{array}\right]$$ 
		and 
		\begin{equation}
	\begin{aligned}
	{\rqq}(x,y)&=\Re_c^{-1}([r_0(x,y), r_1(x,y), r_2(x,y),r_3(x,y)]^T)\\
		&=r_0(x,y)+r_1(x,y)\iq+r_2(x,y)\jq+r_3(x,y)\kq.
			\end{aligned}
\end{equation}
Then
\begin{equation}
	\begin{aligned}
	B\star\Re_c(\uq(x,y))&=Q\star\Re_c(\uq(x,y))+R\star\Re_c(\uq(x,y))\\
	&=\Re(\Qq)\star\Re_c(\uq(x,y)+\Re_c(\rqq(x,y))\\
	&=\Re_c(\Qq \textcircled{$\star$} \uq(x,y)+\rqq(x,y)).
	\end{aligned}
\end{equation}

Recall the definition of  transformation map  $\mathcal{T}$ in the main body that
 \begin{equation}\label{Tfuna}
 \mathcal{T}(\qq(x,y)):=q(x,y)=[q_1(x,y), q_2(x,y),q_3(x,y)]^T
 \end{equation}
 for $\qq(x,y)=q_0(x,y)+q_1(x,y)\iq+q_2(x,y)\jq+q_3(x,y)\kq$.
The blurring process  \eqref{e:expandUa} can be rewritten as
\begin{equation}\label{e:blurmodel0qa}
	\hat{u}(x,y)= \hat{K} \star u(x,y)= \mathcal{T}(\Qq \textcircled{$\star$} \uq(x,y)+\rqq(x,y)),
\end{equation}
where $\Qq=\Re^{-1}(Q)$ and $\rqq(x,y)=\Re_c^{-1}(R\star \Re_c(\uq(x,y)))$.

Moreover, if $B_{ij}$'s in \eqref{e:Q0123a} are set zero operators then we obtain the special  quaternion operator $\Qq$ in Theorem 1 of the main body. 
\end{proof}

In the discrete form, the cross-channel blurring process is in fact a sum of a quaternion matrix-vector product $\Qq\uq$ and a quaternion vector $ \rqq$.  The quaternion operations can preserve the ratio of red, green and blue information of  color pixels in the blurring process and thus,  color confusion and interinfection can be reduced in the deblurring process.  As a result, the correlation of color channels are well preserved and the  recovered color images can achieve a high quality. 
 In theory, there always exists an invertible quaternion matrix $\Qq$ such that $\Qq\uq=\widehat{\uq}$. 
However, it is impractical  to compute such quaternion matrix,  since  a large number of pairs of  original and observed color images under the same blurring process are in need.  Especially,  the original color image is definitely unknown. 
So we alternatively  construct $\Qq$ according to \eqref{e:Qmatrixa}, \eqref{e:Q0123a} and \eqref{e:Bminimizationa} to minimize $\|\rqq\|_2$  in this paper. 

The blocks $B_{ij}$'s of matrix $B$ defined by \eqref{e:Bmatrixa} can be chosen as the solution of the minimization problem:
\begin{equation}\label{e:Bminimizationa}
\begin{aligned}
		&\mathop{\min}\limits_{B_{ij}}~\|B-Q\|_F\\
		&\text{s.t.}~~B_{12}\star u_1+B_{1,3}\star u_2+B_{14}\star u_3=0.
\end{aligned}
\end{equation}
In this way, the quaternion function $\rqq(x,y)$ in \eqref{e:blurmodel0qa} is minimized.

\subsection{The Euler-Lagrange equation of CSTV regularization model}
In this section, we derive  the Euler-Lagrange equation of CSTV regularization model. Without causing any confusion, we will use $\uq(x,y)$ to represent  the quaternion function $u_1(x,y)\iq+u_2(x,y)\jq+u_3(x,y)\kq$ or the real vector function 
$[u_1(x,y),u_2(x,y),u_3(x,y)]^T$ in the following text. Sometimes, $(x,y)$ will ignored for short. 

We still explore $p=2$ as an example in terms of its corresponding Euler-Lagrange equation.
The Euler-Lagrange equation corresponding to the function of several variables  can be expressed as follows,
\begin{equation}\label{e:ori_ELa}
	\frac{\partial f}{\partial u}-\frac{\partial}{\partial x} \frac{\partial f}{\partial_x u}-\frac{\partial}{\partial y} \frac{\partial f}{\partial_y u}=0.
\end{equation}
If expressed by the divergence operator, $\eqref{e:ori_ELa}$ can be equivalently written as
\begin{equation}
	\nabla \cdot \left(\frac{\partial f}{\partial_x u}+\frac{\partial f}{\partial_y u}\right) - \frac{\partial f}{\partial u}=0.
\end{equation}
To facilitate the representation, give the following notation,
\begin{equation}
	\begin{aligned}
		f_1&:=\sqrt{(|\partial_x\uq(x,y)|_s^2)+(|\partial_y\uq(x,y)|_s^2)}\\ &+\alpha\sqrt{(|\partial_x\uq(x,y)|_v^2)+(|\partial_y\uq(x,y)|_v^2)},\\
		f_2&:=\sqrt{\sum_{i=1}^{3}(\partial_xu_i(x,y))^2+(\partial_yu_i(x,y))^2},\\
		f_3&:=\frac{1}{2}|\Kq\star\uq-\zq|^2.
	\end{aligned}
\end{equation}
The Euler-Lagrange equation is constructed with the above notation. Applying the partial derivative to $f_1$, we have 
\begin{equation}\label{e:par_s2a}
	|\partial_x \uq|_s^2=\frac{1}{9}(\partial_x \uq)^{T}C^{T}C(\partial_x \uq),
\end{equation}
and
\begin{equation}
	\left(\left|\partial_x \uq\right|_s^2\right)_{\partial_x \uq}^{\prime}=\frac{1}{9} C^{T}C\left(\partial_x \uq\right)=\partial_x (C^2\uq).
\end{equation}
The above equation can be further simplified as
$$\left(|\partial_x \uq|_s^2\right)_{\partial_x\uq}^{\prime}=\frac{1}{3}\partial_x(C\uq).$$
A similar result can be obtained for
$$\left(\left|\partial_y\uq\right|_s^2\right)_{\partial_y\uq}^{\prime}=\frac{1}{3}\partial_y(C\uq).$$
For $| \cdot|_v$, we have
\begin{equation}\label{e:par_v2a}
	|\partial_x \uq|_v^2=\frac{1}{3}(\partial_x\uq)^{T}(\partial_x\uq),
\end{equation}
and
\begin{equation}
	\left(\left|\partial_x \uq\right|_v^2\right)_{\partial_x\uq}^{\prime}=\frac{1}{3}(\partial_x\uq).
\end{equation}
A similar result can be obtained for
$$\left(\left|\partial_y\uq\right|_v^2\right)_{\partial_y\uq}^{\prime}=\frac{1}{3}(\partial_y\uq).$$
Therefore, for $f_1$, it holds as follows
\begin{equation}\label{e:parf1_1a}
	\begin{aligned}
		\frac{\partial f_1}{\partial_x \uq}=\frac{1}{3}\left(\frac{\partial_x C\uq}{\sqrt{| \partial_x \uq|_s^2+|\partial_y \uq|_s^2}}+\frac{\partial _x \uq}{\sqrt{|\partial_x \uq|_v ^2+\left|\partial_y \uq\right|_v^2}}\right),\\
		\frac{\partial f_1}{\partial_y \uq}=\frac{1}{3}\left(\frac{\partial_y C\uq}{\sqrt{| \partial_x \uq|_s^2+|\partial_y \uq|_s^2}}+\frac{\partial _y \uq}{\sqrt{|\partial_x \uq|_v ^2+\left|\partial_y \uq\right|_v^2}}\right).
	\end{aligned}
\end{equation}
In addition, by the definition of $f_1$ it follows that
\begin{equation}\label{e:parf1_2a}
	\frac{\partial f_1}{\partial \uq} = 0.
\end{equation}
Similarly, $f_2$ and $f_3$ are calculated: applying the partial derivative to  $f_2$ yields
\begin{equation}\label{e:parf2a}
	\begin{aligned}
		\frac{\partial f_2}{\partial_x \uq}=\frac{\partial_x \uq}{\|\nabla\uq(x,y)\|_2},~~
		\frac{\partial f_2}{\partial_y \uq}=\frac{\partial_y \uq}{\|\nabla\uq(x,y)\|_2},~~
		\frac{\partial f_2}{\partial \uq} = 0.
	\end{aligned}
\end{equation}
Applying the partial derivative to  $f_3$ yields
\begin{equation}\label{e:parf3a}
	\begin{aligned}
		\frac{\partial f_3}{\partial_x \uq}=0,~~
		\frac{\partial f_3}{\partial_y \uq}=0,~~
		\frac{\partial f_3}{\partial \uq} = (\Kq\star \uq-\zq)\star\Kq^{*},
	\end{aligned}
\end{equation}
where $\Kq^{*}$ is the conjugate transpose of $\Kq$. Combining the above $\eqref{e:parf1_1a}$~,~$\eqref{e:parf1_2a}$~,~$\eqref{e:parf2a}$~,~$\eqref{e:parf3a}$, one arrives at the Euler-Lagrange equation in the following form
\begin{equation}\label{ELeqa}
	\begin{aligned}[c]
		&\frac{\lambda_1}{3} \nabla\cdot \Bigg( \frac{\nabla\left(\uq C  \right) }{\sqrt{(|\partial_x\uq(x,y)|_s^2)+(|\partial_y\uq(x,y)|_s^2)}}\\
		&+\alpha\frac{\nabla\uq  }{\sqrt{(|\partial_x\uq(x,y)|_v^2)+(|\partial_y\uq(x,y)|_v^2)}} \Bigg)\\
		&+\lambda_2\frac{\nabla \uq}{\|\nabla \uq\|_2}
		-\left( \Kq\star \uq-\star \zq \right)\star \Kq^{*} =0.	
	\end{aligned}
\end{equation} 
In \eqref{ELeqa}, the proposed model takes into account both the diffusion coefficients of the channel coupling over the saturation and value components and the diffusion coefficients of the summation of the individual channels over the RGB space.
The quaternion representation also leads to many other advantages, such as the values of three color channels of a color pixel are flocked together and their physical meanings are preserved  in the whole color image processing based on quaternion computation.  Especially, color distortion will be hugely reduced in the recovered color image by the new model.

	Certainly, both regular terms constrain the gradient information of the image, and in the following the two regularization terms are considered as a whole, so that it is more obvious to see how the CSTV  regularization term acts on the gradient. It is sufficient to transform the denominator of the SVTV term. According to the above derivations, the denominator of the SVTV term can be reduced to
	\begin{equation}
		\begin{aligned}
			\sqrt{| \partial_x \uq|_s^2+|\partial_y \uq|_s^2}&=\frac{1}{3}\sqrt{(\nabla C\uq)^{T}(\nabla C\uq)}=\frac{1}{3}\|\nabla C\uq\|_2,\\
			\sqrt{|\partial_x \uq|_v ^2+\left|\partial_y \uq\right|_v^2}&=\sqrt{\frac{1}{3}(\nabla\uq)^{T}(\nabla\uq)}=\frac{\sqrt{3}}{3}\|\nabla\uq\|_2.
		\end{aligned}
	\end{equation}
It is interesting to note that, without considering the coefficients, the form of the contribution of the V-component to the gradient in the SVTV  is identical to the form of the contribution of the CTV to the gradient.
Considering the symmetry of the matrix $C$, \eqref{ELeqa} can be rewritten as
\begin{equation}\label{e:EL_wholea}
	\begin{aligned}
		&\nabla \cdot(\lambda_1\frac{\nabla  \frac{C}{3}\uq}{\|(\nabla  \frac{C}{3}\uq)\|_2}+(\frac{\sqrt{3}}{3}\alpha\lambda_1+\lambda_2)\frac{\nabla \uq}{\|(\nabla\uq)\|_2})\\
		&- (\Kq\star  \uq-\zq)\star \Kq^{*}=0.
	\end{aligned}
\end{equation}
From equation \eqref{e:EL_wholea}, it is easy to see that the CSTV regularization model can be viewed as a weighted combination of S and V components in HSV color space to achieve the effect of restoring the color image using both HSV and RGB color space information. This is essentially explained by the fact that the R, G, and B components are all expressed in terms of value, i.e., the V component. Note that it is  impractical to solve the Euler-Lagrange equation  \eqref{e:EL_wholea} directly, since  a non-linear differential equation  leads to a non-linear system after discretization.

Next, we present the dual form of  the CSTV regularization model.
The dual forms of regularization functions SVTV and CTV are
	\begin{equation*}
		\begin{aligned}
		\SVTV(\uq)&=\max_{\|g\|\leq 1}~\left<\diag(1,1,\alpha)\Pq\uq,\nabla g\right>,\\
		\CTV(\uq)&=\max_{\|h\|\leq 1}~\left<\uq,\nabla h\right>,
		\end{aligned}
	\end{equation*}
where $\Pq$ is an orthogonal transformation matrix that takes the form of
	\begin{equation*}
		\Pq=\left [\begin{array}{ccc}
			\frac{1}{\sqrt{2}}\Iq&\frac{-1}{\sqrt{2}}\Iq& 0\\
			\frac{1}{\sqrt{6}}\Iq&\frac{1}{\sqrt{6}}\Iq&\frac{-2}{\sqrt{6}}\Iq\\
			\frac{1}{\sqrt{3}}\Iq&\frac{1}{\sqrt{3}}\Iq&\frac{1}{\sqrt{3}}\Iq
		\end{array}\right ].
	\end{equation*}
	By converting the regularization terms into dual forms, we can rephrase the CSTV regularization model as the following dual form:
	\begin{equation}\label{m:dualCSmodela}
		\min\limits_{\uq}\!\max\limits_{\|g\|\leq 1
			\atop
			\|h\|\leq 1}\lambda_1\!\left<\diag(1,1,\alpha)\Pq\uq,\nabla g\right>+\lambda_2\!\left<\uq,\nabla h\right>+
 \atop
\frac{1}{2}\|\Qq\textcircled{$\star$}\uq+\rqq-\zq\|^2.
	\end{equation}
In the dual form, the solution $\uq(x,y)$ can be differentiable or not, which expands the feasible set.   Obviously, the min-max problem \eqref{m:dualCSmodela} is a new saddle-point problem.

There are various methods to solve the CSTV regularization model or its dual form \eqref{m:dualCSmodela}  with real variables. However, there no methods for solving  them  with quaternion variables.

\subsection{Augmented Lagrangian method for $(\uq,\vq)$-subproblem} \label{ss:wqvqsubproblema}
Now, we present a new augmented Lagrangian method for solving $(\wq,\vq)$-subproblem,
	\begin{equation}\label{model_wva}
		\mathop{\arg\min}\limits_{\wq,\vq}~\lambda_1\SVTV(\wq)+\frac{\alpha_1}{2}\|\wq-\uq\|^2+\atop \lambda_2\CTV(\vq)+ \frac{\alpha_2}{2}\|\vq-\uq\|^2.
	\end{equation}
  Since the two variables are independent of each other and have no intersecting terms, we construct the augmented Lagrangian schemes of computing $\wq$ and $\vq$, respectively.

\begin{itemize}	
	\item[$\bullet$]$\wq$-subproblem
	
	With $\uq$ and $\vq$ fixed,   the minimization problem \eqref{model_wva} is reduced to
	\begin{equation}\label{model_wa}
		\mathop{\arg\min}\limits_{\wq}~\lambda_1\SVTV(\wq)+\frac{\alpha_1}{2}\|\wq-\uq\|^2.
	\end{equation}
	
	We define two discrete differential operators $\Dq_x$, $\Dq_y$,
	\begin{equation*}
		\begin{aligned}
			(\Dq_x\wq)_{i,j}=\wq(i,j)-\wq(i-1,j),\\
			(\Dq_y\wq)_{i,j}=\wq(i,j)-\wq(i,j-1),
		\end{aligned}
	\end{equation*}
under adequate boundary conditions for color image.
	Define 
	\begin{equation*}
		\sq=\Pq\wq
~\text{and}~\qq=\Pq\uq.
	\end{equation*}
	Then the objective color image restoration model \eqref{model_wa} is reformulated as
	\begin{equation}\label{model_sa}
		\begin{aligned}
\mathop{\arg\min}\limits_{\sq}~\frac{\lambda_1}{\alpha_1}\sum\limits_{i=1}^m\sum\limits_{j=1}^n(\sqrt{\sum\limits_{k=1}^2((\Dq_xs_k)_{ij})^2\!+\!((\Dq_ys_k)_{ij})^2}\\
			+\alpha\sqrt{((\Dq_xs_3)_{ij})^2+((\Dq_ys_3)_{ij})^2})+
			\frac{1}{2}\|\sq-\qq\|^2.
		\end{aligned}
	\end{equation} 
 By introducing  auxiliary variables $t_i^x$  and  $t_i^y$, the minimization problem \eqref{model_sa}   is equivalently rewritten into 
	\begin{equation*}
		\begin{aligned}
&\mathop{\arg\min}\limits_{\sq}~\frac{\lambda_1}{\alpha_1}\sum\limits_{i=1}^m\sum\limits_{j=1}^n(\sqrt{\sum\limits_{k=1}^2|(t_k^x)_{ij}|^2+|(t_k^y)_{ij}|^2}\\ 
			&+\alpha\sqrt{|(t_3^x)_{ij}|^2+|(t_3^y)_{ij}^2|})+
			\frac{1}{2}\|\sq-\qq\|^2,\\
			\text{s.t.}~~&t_1^x=\Dq_xs_1,t_2^x=\Dq_xs_2,t_3^x=\Dq_xs_3,\\
			&t_1^y=\Dq_ys_1,t_2^y=\Dq_ys_2,t_3^y=\Dq_ys_3.
		\end{aligned}
	\end{equation*}
	The augmented Lagrangian of the aforementioned minimization problem is
	\begin{equation}\label{alma}
		\begin{aligned}
			&\frac{\lambda_1}{\alpha_1}\sum\limits_{i=1}^m\sum\limits_{j=1}^n(\sqrt{\sum\limits_{k=1}^2|(t_k^x)_{ij}|^2+|(t_k^y)_{ij}|^2}\\
			&+\alpha\sqrt{|(t_3^x)_{ij}|^2+|(t_3^y)_{ij}^2|})+\frac{1}{2}\|\sq-\qq\|^2\\
			&+\sum\limits_{i=1}^3((\tau_i^x,t_i^x-\Dq_xs_i)+(\tau_i^y,t_i^y-\Dq_ys_i))\\
			&+\frac{\beta}{2}\sum\limits_{i=1}^3(\|t_i^x-\Dq_xs_i\|^2+\|t_i^y-\Dq_ys_i\|^2),
		\end{aligned}
	\end{equation}
where  $\tau_i^x$ and $\tau_i^y$ are  Lagrangian multipliers and  $\beta$ is a positive penalty parameter.
	
	According to the classic framework of the ADMM \cite{e09a,jd92a}, we construct a method to sovle  \eqref{alma} with three steps as follows.
	
	{\bf Step one: } When fixing $\sq$,  $t_i^{x}$  and $t_i^{y}$ are computed by using the soft shrinkage  functions,
	\begin{subequations}\label{t-subpba}
		\begin{align}
			t_i^{x,y}=&\max{(0,n_1-\frac{\lambda_1}{\beta\alpha_1})}\cdot\frac{\Dq_xs_i-\frac{\tau_i^{x,y}}{\beta}}{n_1},~~~~i=1,2,\\
			t_3^{x,y}=&\max{(0,n_2-\frac{\alpha\lambda_1}{\beta\alpha_1})}\cdot\frac{\Dq_xs_3-\frac{\tau_3^{x,y}}{\beta}}{n_2},			
		\end{align}
	\end{subequations}
	where
	\begin{equation*}
		\begin{aligned}
			n_1&=\sqrt{\sum\limits_{i=1}^2(\Dq_xs_i-\frac{\tau_i^x}{\beta})^2+(\Dq_ys_i-\frac{\tau_i^y}{\beta})^2},\\
			n_2&=\sqrt{(\Dq_xs_3-\frac{\tau_3^x}{\beta})^2+(\Dq_ys_3-\frac{\tau_3^y}{\beta})^2}.
		\end{aligned}
	\end{equation*}
	{\bf Step two: } When fixing $t_i^x$ and $t_i^y$,  $\sq$ is computed by solving the following  minimization problem
	\begin{equation*}
		\begin{aligned}
			\frac{1}{2}\|\sq-\qq\|^2+\sum\limits_{i=1}^3((\tau_i^x,t_i^x-\Dq_xs_i)+(\tau_i^y,t_i^y-\Dq_ys_i))\\
			+\frac{\beta}{2}\sum\limits_{i=1}^3(\|t_i^x-\Dq_xs_i\|^2+\|t_i^y-\Dq_ys_i\|^2).
		\end{aligned}
	\end{equation*}
	The above minimization problem can be equivalently described by the following real linear system
	\begin{equation}\label{s-subpba}
		\begin{aligned}
			(\Iq+\beta\Dq_x^T&\Dq_x+\beta\Dq_y^T\Dq_y)\sq\\
			=\qq+\sum\limits_{i=1}^3(\Dq_x^T\tau_i^x+&\Dq_y^T\tau_i^y+\beta\Dq_x^Tt_i^x+\beta\Dq_y^Tt_i^y).
		\end{aligned}
	\end{equation}
Once $\sq$ is computed, we obtain $\wq=\Pq^{-1}\sq$.
	
	{\bf Step three: } The Lagrangian multipliers are updated as follows
	$$\tau_i^x \!=\! \tau_i^x+\beta(t_i^x-\Dq_xs_i),~~\tau_i^y \!=\! \tau_i^y+\beta(t_i^y-\Dq_ys_i),~i=1,2,3.$$

	\item [$\bullet$]$\vq$-subproblem
	
	With  $\uq$ an $\wq$ fixed,   the minimization problem \eqref{model_wva} is reduced to
	\begin{equation}\label{model_va}
		\mathop{\arg\min}\limits_{\vq}~\lambda_2\CTV(\vq)+\frac{\alpha_2}{2}\|\vq-\uq\|^2.
	\end{equation}

We define two discrete differential operators $\Dq_x$, $\Dq_y$,
\begin{equation*}
	\begin{aligned}
		(\Dq_x\vq)_{i,j}=\vq(i,j)-\vq(i-1,j),\\
		(\Dq_y\vq)_{i,j}=\vq(i,j)-\vq(i,j-1),
	\end{aligned}
\end{equation*}
under adequate boundary conditions for color image.
Then the objective color image restoration model \eqref{model_va} is reformulated as
\begin{equation}\label{model_v2a}
	\begin{aligned}		\vq^*\!=\!\mathop{\arg\min}\limits_{\vq}~&\frac{\lambda_2}{\alpha_2}\sum\limits_{i=1}^m\sum\limits_{j=1}^n(\sqrt{\sum\limits_{k=1}^3((\Dq_xv_k)_{ij})^2\!+\!((\Dq_yv_k)_{ij})^2}\\
	&+\frac{1}{2}\|\vq-\uq\|^2.
	\end{aligned}
\end{equation}
By introducing  auxiliary variables $l_i^x$  and  $l_i^y$, the minimization problem \eqref{model_v2a}   is equivalently rewritten into 
\begin{equation*}
	\begin{aligned}		\vq^*=&\mathop{\arg\min}\limits_{\vq}~\frac{\lambda_2}{\alpha_2}\sum\limits_{i=1}^m\sum\limits_{j=1}^n(\sqrt{\sum\limits_{k=1}^3|(l_k^x)_{ij}|^2+|(l_k^y)_{ij}|^2}\\
	&+\frac{1}{2}\|\vq-\uq\|^2,\\		\text{s.t.}~~&l_1^x=\Dq_xv_1,l_2^x=\Dq_xv_2,l_3^x=\Dq_xv_3,\\
		&l_1^y=\Dq_yv_1,l_2^y=\Dq_yv_2,l_3^y=\Dq_yv_3.
	\end{aligned}
\end{equation*}
The augmented Lagrangian of the aforementioned minimization problem is
\begin{equation}\label{alm_va}
	\begin{aligned}		&\frac{\lambda_2}{\alpha_2}\sum\limits_{i=1}^m\sum\limits_{j=1}^n(\sqrt{\sum\limits_{k=1}^3|(l_k^x)_{ij}|^2+|(l_k^y)_{ij}|^2}+\frac{1}{2}\|\vq-\uq\|^2\\ &+\sum\limits_{i=1}^3((\eta_i^x,l_i^x-\Dq_xv_i)+(\eta_i^y,l_i^y-\Dq_yv_i))\\ &+\frac{\beta_2}{2}\sum\limits_{i=1}^3(\|l_i^x-\Dq_xv_i\|^2+\|l_i^y-\Dq_yv_i\|^2).
	\end{aligned}
\end{equation}
where  $\eta_i^x$ and $\eta_i^y$ are  Lagrangian multipliers and  $\beta_2$ is a positive penalty parameter.
According to the classic framework of the ADMM \cite{e09a,jd92a}, we construct a method to sovle  \eqref{alm_va} with three steps as follows.
{\bf Step one: } When fixing $\vq$,  $l_i^{x}$  and $l_i^{y}$ are computed by using the soft shrinkage  functions,
\begin{equation}\label{e:l-subpba}	l_i^{x,y}=\max{(0,m-\frac{\lambda_2}{\beta_2\alpha_2})}\cdot\frac{\Dq_xv_i-\frac{\eta_i^{x,y}}{\beta_2}}{m},~~~~i=1,2,3,
\end{equation}
where
\begin{equation*}
	m=\sqrt{\sum\limits_{i=1}^3(\Dq_xv_i-\frac{\eta_i^x}{\beta_2})^2+(\Dq_yv_i-\frac{\eta_i^y}{\beta_2})^2}.
\end{equation*}
{\bf Step two: } When fixing $l_i^x$ and $l_i^y$,  $\vq$ is computed by solving the following  minimization problem
\begin{equation*}
	\begin{aligned}		\frac{1}{2}\|\vq-\uq\|^2+\sum\limits_{i=1}^3((\eta_i^x,l_i^x-\Dq_xv_i)+(\eta_i^y,l_i^y-\Dq_yv_i))\\ +\frac{\beta_2}{2}\sum\limits_{i=1}^3(\|l_i^x-\Dq_xv_i\|^2+\|l_i^y-\Dq_yv_i\|^2).
	\end{aligned}
\end{equation*}
The above minimization problem can be equivalently described by the following real linear system
\begin{equation}\label{v-subpba}
	\begin{aligned}	(\Iq+\beta_2\Dq_x^T&\Dq_x+\beta_2\Dq_y^T\Dq_y)\vq\\ =\uq+\sum\limits_{i=1}^3(\Dq_x^T\eta_i^x+&\Dq_y^T\eta_i^y+\beta_2\Dq_x^Tl_i^x+\beta_2\Dq_y^Tl_i^y).
	\end{aligned}
\end{equation}
{\bf Step three: } The Lagrangian multipliers are updated as follows
\begin{equation*}
	\begin{aligned}
		 &\eta_i^x \!=\! \eta_i^x+\beta_2(l_i^x-\Dq_xv_i),\\
		 &\eta_i^y \!=\! \eta_i^y+\beta_2(l_i^y-\Dq_yv_i)~~(i=1,2,3).
	\end{aligned}	 
\end{equation*}
\end{itemize}

In summary, the $(\wq,\vq)$-subproblem needs to compute several soft shrinkage functions and solve two real linear systems.

In this paper, we choose to determine  $[\lambda_1,\lambda_2] $ first and subsequently optimize $[\alpha_1,\alpha_2]$ under the determined optimal  $[\lambda_1,\lambda_2] $ parameters setting. This parameter selection strategy is formulated based on the following considerations:  $[\lambda_1,\lambda_2] $, as the regularization parameters, have more significant impact on the image restoration results. Therefore prioritizing  $[\lambda_1,\lambda_2] $ can help stabilize the model performance at an early stage and ensure that the subsequent optimization of $[\alpha_1,\alpha_2]$ parameters can be carried out on a more optimal basis, leading to more accurate restoration results.

Theoretically, the selection of  $[\lambda_1,\lambda_2] $  is related to the noise level.  In most of  numerical experiments, we did not adjust  $[\lambda_1,\lambda_2] $  according to changes of noise level, but rather fixed $ \sigma $ at 0.01. The corresponding parameter pairs were initially selected using the L-surface method and then refined through further tuning. This approach effectively reduces the difficulty and time cost of parameter tuning. 
	
	If the noise level changes, the optimal  $[\lambda_1,\lambda_2] $  parameter pair would also change accordingly.  We test this case at the end of  Section  V-B of the main body. The values of model parameters are given later in Table I  of the supplementary material.  Adapting the selection of multiple coupled regularization parameters based on the noise level will be a part of our future work.

\subsection{Real algorithms of solving the CSTV regularization model}
In this section, we present  a real algorithm of solving the CSTV regularization model in the main body, which is denoted by CSTV$_{\rm GMRES}$.

Following the real ADMM method, we need solve three subproblems alternatively. 
\begin{itemize}
	\item[$\bullet$]$\uq$-subproblem
	
	Fixing $\wq$ and $\vq$ simplifies the minimization problem to
	\begin{equation}\label{model_ua}
		\mathop{\arg\min}\limits_{\uq}~	\frac{1}{2}\|\Kq\star\uq-\zq\|^2+\frac{\alpha_1}{2}\|\wq-\uq\|^2+\frac{\alpha_2}{2}\|\vq-\uq\|^2.
	\end{equation}
	This is essentially a least squares problem, which is the same as solving
	\begin{equation}\label{ls_ua}
		(\Kq^T\Kq+\alpha_1\Iq+\alpha_2\Iq)\uq=\Kq^T\zq+\alpha_1\wq+\alpha_2\vq.
	\end{equation}
	In this case $\Kq$ is a real matrix, so it can be solved by the general numerical method for the solution of the real linear system. In this paper, we use the real generalized minimal residual method (GMRES), which reduces to the Lanczos method when the coefficient matrix is symmetric \cite{saad03a}. 
	
	\item[$\bullet$]$\wq$-subproblem
	
	Fixing $\uq$ simplifies the minimization problem to
	\begin{equation}\label{model_wa}
		\mathop{\arg\min}\limits_{\wq}~\frac{\lambda_1}{\alpha_1}\SVTV(\wq)+\frac{1}{2}\|\wq-\uq\|^2.
	\end{equation}
	
	We define two discrete differential operators $\Dq_x$, $\Dq_y$,
	\begin{equation*}
		\begin{aligned}
			(\Dq_x\wq)_{i,j}=\wq(i,j)-\wq(i-1,j),\\
			(\Dq_y\wq)_{i,j}=\wq(i,j)-\wq(i,j-1).
		\end{aligned}
	\end{equation*}
	Always assume adequate boundary conditions for color image.
	To better solve the minimization problem \eqref{model_wa}, the linear transformation $\Pq$ is applied to $\wq$, 
	and
	\begin{equation*}
		\sq=\left [\begin{array}{c}
			s_1,
			s_2,
			s_3
		\end{array}\right ]^T=\Pq\left[
		\begin{array}{c}
			w_r,
			w_g,
			w_b
		\end{array}\right]^T.
	\end{equation*}
	Without failing the general, we set $\qq=\Pq\uq$.
	Then the objective color image restoration model \eqref{model_wa} is reformulated as
	\begin{equation}\label{model_sa}
		\begin{aligned}
			\sq^*\!=\!\mathop{\arg\min}\limits_{\sq}~\frac{\lambda_1}{\alpha_1}\sum\limits_{i=1}^m\sum\limits_{j=1}^n(\sqrt{\sum\limits_{k=1}^2((\Dq_xs_k)_{ij})^2\!+\!((\Dq_ys_k)_{ij})^2}\\
			+\alpha\sqrt{((\Dq_xs_3)_{ij})^2+((\Dq_ys_3)_{ij})^2})+
			\frac{1}{2}\|\sq-\qq\|^2.
		\end{aligned}
	\end{equation} 
	
	By introducing auxiliary variables, the minimization problem \eqref{model_sa} can be rewritten as an equivalent minimization problem,
	\begin{equation*}
		\begin{aligned}
			\sq^*=&\mathop{\arg\min}\limits_{\sq}~\frac{\lambda_1}{\alpha_1}\sum\limits_{i=1}^m\sum\limits_{j=1}^n(\sqrt{\sum\limits_{k=1}^2|(t_k^x)_{ij}|^2+|(t_k^y)_{ij}|^2}\\ 
			&+\alpha\sqrt{|(t_3^x)_{ij}|^2+|(t_3^y)_{ij}^2|})+
			\frac{1}{2}\|\sq-\qq\|^2,\\
			\text{s.t.}~~&t_1^x=\Dq_xs_1,t_2^x=\Dq_xs_2,t_3^x=\Dq_xs_3,\\
			&t_1^y=\Dq_ys_1,t_2^y=\Dq_ys_2,t_3^y=\Dq_ys_3.
		\end{aligned}
	\end{equation*}
	The augmented Lagrangian of the aforementioned minimization problem is as follows after introducing Lagrangian multipliers $\tau_i^x$ , $\tau_i^y$ $(i=1,2,3)$ and the positive penalty parameter $\beta$,
	\begin{equation}\label{alma2}
		\begin{aligned}
			&\frac{\lambda_1}{\alpha_1}\sum\limits_{i=1}^m\sum\limits_{j=1}^n(\sqrt{\sum\limits_{k=1}^2|(t_k^x)_{ij}|^2+|(t_k^y)_{ij}|^2}\\
			&+\alpha\sqrt{|(t_3^x)_{ij}|^2+|(t_3^y)_{ij}^2|})+\frac{1}{2}\|\sq-\qq\|^2\\
			&+\sum\limits_{i=1}^3((\tau_i^x,t_i^x-\Dq_xs_i)+(\tau_i^y,t_i^y-\Dq_ys_i))\\
			&+\frac{\beta}{2}\sum\limits_{i=1}^3(\|t_i^x-\Dq_xs_i\|^2+\|t_i^y-\Dq_ys_i\|^2).
		\end{aligned}
	\end{equation}
	
	The minimum score of \eqref{alma2} can be achieved by using the classic framework of the ADMM \cite{e09a,jd92a}.
	
	When fixing $\sq$, the solutions are giving by using the soft shrinkage algorithm,
	\begin{subequations}\label{t-subpba}
		\begin{align}
			t_i^{x,y}=&\max{(0,n_1-\frac{\lambda_1}{\beta\alpha_1})}\cdot\frac{\Dq_xs_i-\frac{\tau_i^{x,y}}{\beta}}{n_1},~~~~i=1,2,\\
			t_3^{x,y}=&\max{(0,n_2-\frac{\alpha\lambda_1}{\beta\alpha_1})}\cdot\frac{\Dq_xs_3-\frac{\tau_3^{x,y}}{\beta}}{n_2},			
		\end{align}
	\end{subequations}
	where
	\begin{equation*}
		\begin{aligned}
			n_1&=\sqrt{\sum\limits_{i=1}^2(\Dq_xs_i-\frac{\tau_i^x}{\beta})^2+(\Dq_ys_i-\frac{\tau_i^y}{\beta})^2},\\
			n_2&=\sqrt{(\Dq_xs_3-\frac{\tau_3^x}{\beta})^2+(\Dq_ys_3-\frac{\tau_3^y}{\beta})^2}.
		\end{aligned}
	\end{equation*}
	When fixing $t_i^x, t_i^y$, the minimization problem is reduced to
	\begin{equation*}
		\begin{aligned}
			\frac{1}{2}\|\sq-\qq\|^2+\sum\limits_{i=1}^3((\tau_i^x,t_i^x-\Dq_xs_i)+(\tau_i^y,t_i^y-\Dq_ys_i))\\
			+\frac{\beta}{2}\sum\limits_{i=1}^3(\|t_i^x-\Dq_xs_i\|^2+\|t_i^y-\Dq_ys_i\|^2).
		\end{aligned}
	\end{equation*}
	Then, the above minimization problem can be equivalently described by the following real linear system
	\begin{equation}\label{s-subpba}
		\begin{aligned}
			(\Iq+\beta\Dq_x^T&\Dq_x+\beta\Dq_y^T\Dq_y)\sq\\
			=\qq+\sum\limits_{i=1}^3(\Dq_x^T\tau_i^x+&\Dq_y^T\tau_i^y+\beta\Dq_x^Tt_i^x+\beta\Dq_y^Tt_i^y).
		\end{aligned}
	\end{equation}
	The new value of $\wq$, after satisfying the stopping criteria and exiting the iteration, is $\Pq^{-1}\sq$.
	
	The laws for updating the Lagrangian multipliers are as follows
	$$\tau_i^x \!=\! \tau_i^x+\beta(t_i^x-\Dq_xs_i),~~\tau_i^y \!=\! \tau_i^y+\beta(t_i^y-\Dq_ys_i)(i=1,2,3).$$
	
	\item [$\bullet$]$\vq$-subproblem
	
	Fixing $\uq$ allows $\vq$ to be refreshed at the same time.
	\begin{equation}\label{model_va}
		\mathop{\arg\min}\limits_{\vq}~\frac{\lambda_2}{\alpha_2}\CTV(\vq)+\frac{1}{2}\|\vq-\uq\|^2.
	\end{equation}
	The solution to the minimization problem \eqref{model_va} is given by \cite{brc08a}. To obtain the minimum value point, this can also be solved by using a method similar to that of the $\wq$-subproblem.
	
\end{itemize}

\subsection{Numerical results for color image debluring with different levels of noise }
		
\begin{figure}[!t]
		\centering
		\subfigure{\includegraphics[width=0.06\textwidth]{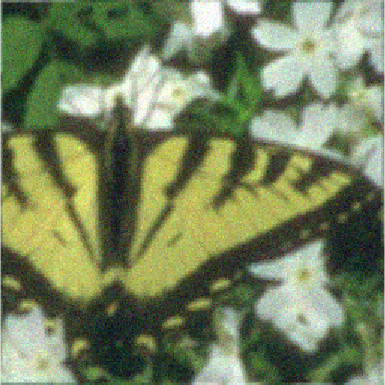}}
		\subfigure{\includegraphics[width=0.06\textwidth]{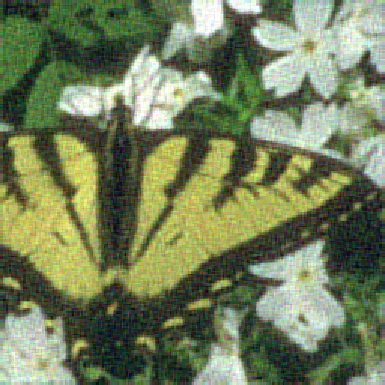}}
		\subfigure{\includegraphics[width=0.06\textwidth]{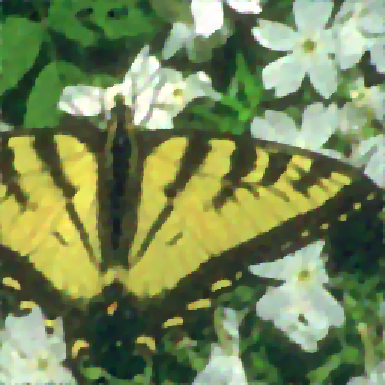}}
		\subfigure{\includegraphics[width=0.06\textwidth]{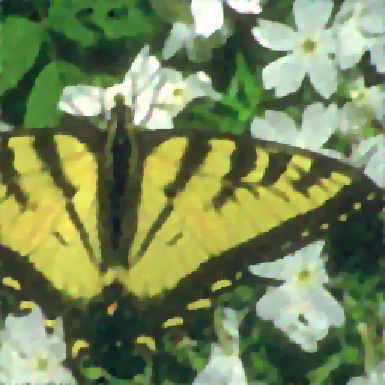}}
		\subfigure{\includegraphics[width=0.06\textwidth]{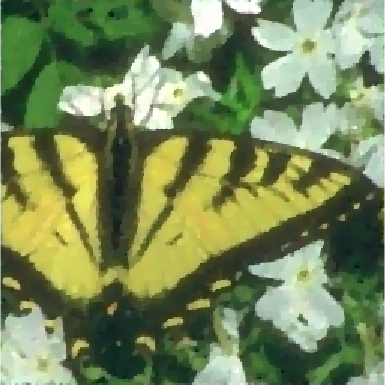}}		
                  \subfigure{\includegraphics[width=0.06\textwidth]{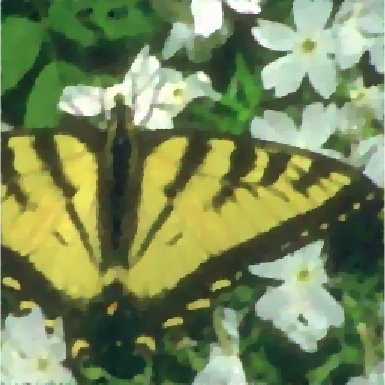}}
		\subfigure{\includegraphics[width=0.06\textwidth]{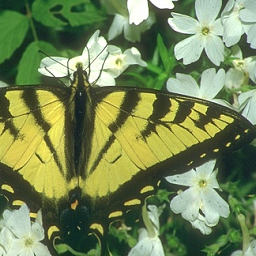}}\\[0.01cm]	
		\subfigure{\includegraphics[width=0.06\textwidth]{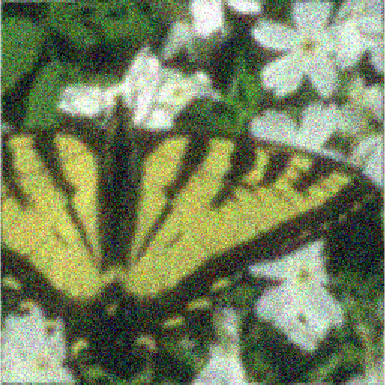}}
		\subfigure{\includegraphics[width=0.06\textwidth]{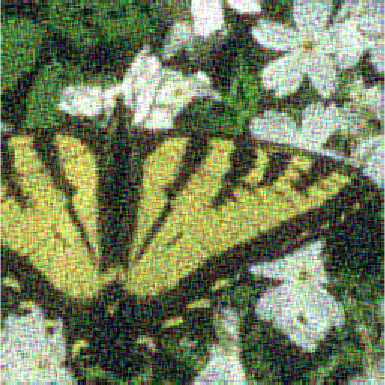}}
		\subfigure{\includegraphics[width=0.06\textwidth]{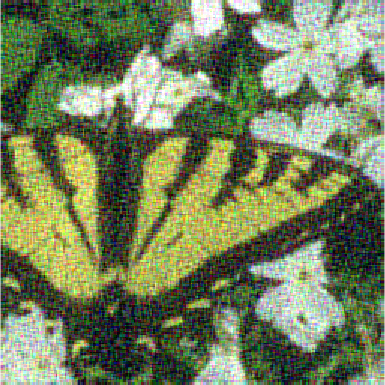}}
		\subfigure{\includegraphics[width=0.06\textwidth]{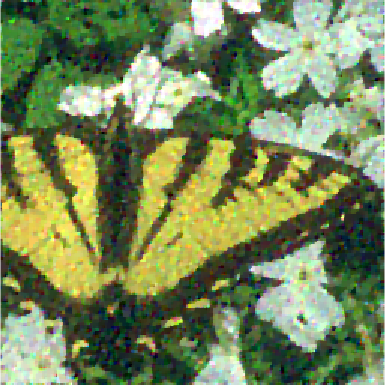}}
		\subfigure{\includegraphics[width=0.06\textwidth]{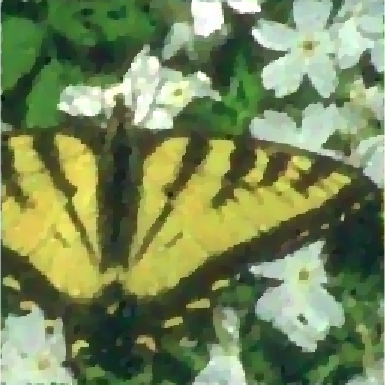}}		
                  \subfigure{\includegraphics[width=0.06\textwidth]{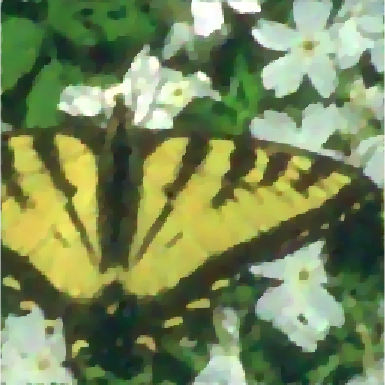}}
		\subfigure{\includegraphics[width=0.06\textwidth]{Figs2/butterfly_ori.png}}\\[0.01cm]	
		\subfigure{\includegraphics[width=0.06\textwidth]{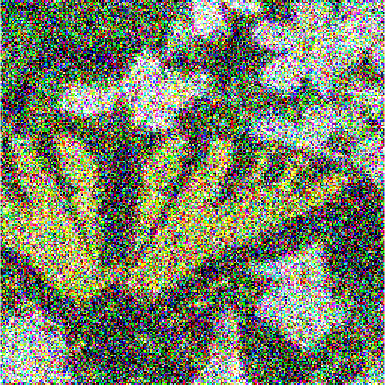}}
		\subfigure{\includegraphics[width=0.06\textwidth]{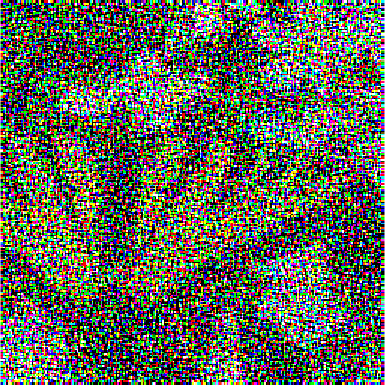}}
		\subfigure{\includegraphics[width=0.06\textwidth]{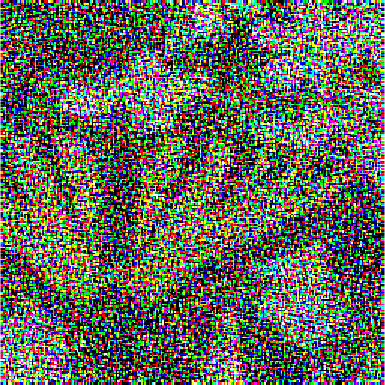}}
		\subfigure{\includegraphics[width=0.06\textwidth]{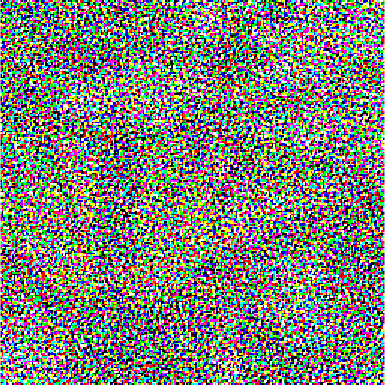}}
		\subfigure{\includegraphics[width=0.06\textwidth]{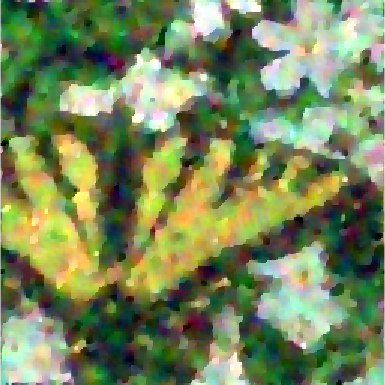}}		
                  \subfigure{\includegraphics[width=0.06\textwidth]{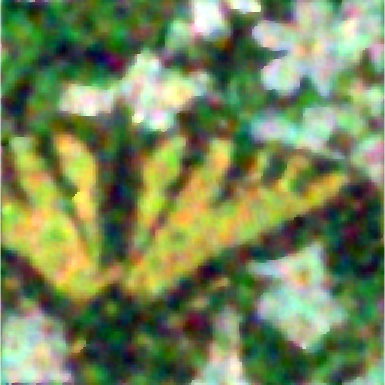}}
		\subfigure{\includegraphics[width=0.06\textwidth]{Figs2/butterfly_ori.png}}\\[0.01cm]	
		(a)\quad~~~ (b)\quad~~~ (c)\quad~~~ (d)\quad~~~ (e)\quad~~~ (f)\quad~~~ (g) 
		\caption{ Visual comparison of five methods in asymmetric cross-channel deblurring corresponding to the numerical results in Table III of Section V-B in the main body. From left to right:  (a) Observed color images,  (b) Restorations of  $\CTV_1$,   (c) Restorations of  $\CTV_2$,   (d) Restorations of   $\MTV$,   (e) Restorations of  $\SVTV$,   (f) Restorations of CSTV,   (g) Original color images.    From up to down, the noise levels are $0.05$, $0.1$ and $0.5$.} 
		\label{fig:modelcom3a}
	\end{figure}

In Figure \ref{fig:modelcom3a}, we present the visual comparison of the compared methods:  $\CTV_1$, $\CTV_2$, $\MTV$, $\SVTV$ and CSTV in the Section V-B of the main body. 
	
In Table \ref{tab:noiselevel3a}, we present the optimal values of parameters of the compared methods  corresponding to the numerical results in Table III of Section V-B in the main body. 

	\begin{table}[htbp]
		\begin{lrbox}{\tablebox}
			\begin{tabular}{|l|c|c|c|c|c|}
				\hline
				Noise level&Method
				&$\lambda_1$&$\alpha_1$ &$\lambda_2$ & $\alpha_2$\\  \hline
\makecell[c]{$0.05$}
				&\makecell[c]{$\CTV_1$\\$\CTV_2$\\$\MTV$\\$\SVTV$\\ $\CSTV$}
				&\makecell[c]{0.1000\\0.0161\\0.0100\\ 0.0443 \\ 0.0398 }
				&\makecell[c]{-\\-\\-\\0.1259\\0.1259}
				&\makecell[c]{-\\-\\-\\-\\0.0028}
				&\makecell[c]{-\\-\\-\\-\\0.1259}
				\\  \hline
\makecell[c]{$0.1$}
				&\makecell[c]{$\CTV_1$\\$\CTV_2$\\$\MTV$\\$\SVTV$\\ $\CSTV$}
				&\makecell[c]{0.1000\\0.0167\\0.0100\\0.1197\\  0.1330 }
				&\makecell[c]{-\\-\\-\\0.2826\\0.2827}
				&\makecell[c]{-\\-\\-\\-\\0.0027}
				&\makecell[c]{-\\-\\-\\-\\0.0426}
				\\  \hline
\makecell[c]{$0.5$}
				&\makecell[c]{$\CTV_1$\\$\CTV_2$\\$\MTV$\\$\SVTV$\\ $\CSTV$}
				&\makecell[c]{0.0909\\ 0.0123\\0.0100\\ 0.7094\\ 0.6651 }
				&\makecell[c]{-\\-\\-\\0.2821\\0.2821}
				&\makecell[c]{-\\-\\-\\-\\0.01995}
				&\makecell[c]{-\\-\\-\\-\\0.0300}
				\\  \hline

			\end{tabular}
		\end{lrbox}
		\centering
		\caption{Optimal values of parameters of the compared methods  corresponding to the numerical results in Table III of Section V-B in the main body}
		\scalebox{0.85}{\usebox{\tablebox}}
		\label{tab:noiselevel3a}
\end{table}

In Table \ref{tab:noiselevel4a}, we present the numbers of iteration steps of the compared methods  corresponding to the setting in Section V-B in the main body. 
\begin{table}[htbp]
		\begin{lrbox}{\tablebox}
			\begin{tabular}{|l|c|c|c|c|}
				\hline
				Method&Outer &Inner  for $\uq$& Inner  for $\wq$ & Inner for $\vq$\\  \hline
				\makecell[c]{$\CTV_1$\\$\CTV_2$\\$\MTV$\\$\SVTV$\\ $\CSTV$}
				&\makecell[c]{3\\3\\35\\ 21 \\ 34 }
				&\makecell[c]{-\\-\\-\\1\\3}
				&\makecell[c]{-\\-\\-\\5\\5}
				&\makecell[c]{-\\-\\-\\-\\5}
				\\  \hline

			\end{tabular}
		\end{lrbox}
		\centering
		\caption{Numbers of iteration steps of the compared method in color image deblurring without noise }
		\scalebox{0.85}{\usebox{\tablebox}}
		\label{tab:noiselevel4a}
\end{table}

\subsection{Injective assumption}
The requirement in Theorem 2 of the main body that the mapping $\uq \mapsto \Qq\textcircled{$\star$}\uq+\rqq$ is injective is motivated by a general assumption for blurring problems. Whether or not mapping $\uq \mapsto \Qq\textcircled{$\star$}\uq+\rqq$ is injective is related to whether or not the matrix corresponding to the cross-channel blurring operator $ \Kq  $ is nonsingular. 
It would be sufficient to show that  $u \mapsto \Kq\star u$ is injective: if $\Kq\star u^{a}=\Kq\star u^{b}$, then $u^{a}=u^{b}$. And under known conditions:
	$$
	\Kq\star
	\left[\begin{array}{rrr}
	u_1^{a}-u_1^{b}\\
	u_2^{a}-u_2^{b}\\
	u_3^{a}-u_3^{b}
	\end{array}\right]=0.$$
	The requirement that the above linear system has only 0 solutions means that the matrix corresponding to $ \Kq $ is required to be nonsingular.

\subsection{HSV color space and quaternion representation}
Although the geometry of HSV color space is different from RGB color space, HSV color space can be represented using quaternion operations. 
	Representing a color image pixel as a pure imaginary quaternion, for any position $ (x, y) $ in the image domain $ \Omega $ we have 
	$$\uq(x,y)=u_1(x,y)\iq+u_2(x,y)\jq+u_3(x,y)\kq.$$
	With the above representation, the components in the HSV color space can be represented as
	\begin{equation*}
		\begin{aligned}
			c_h(x,y)&=\tan^{-1}(\frac{|\uq(x,y)-\bm{\mu}\bm{\nu}\uq(x,y)\bm{\nu}\bm{\mu}|}{\uq(x,y)-\bm{\nu}\uq(x,y)\bm{\nu}}),\\
			c_s(x,y)&=\frac{1}{2}|\uq(x,y)+\bm{\mu}\uq(x,y)\bm{\mu}|,\\
			c_v(x,y)&=\frac{1}{2}|\uq(x,y)-\bm{\mu}\uq(x,y)\bm{\mu}|,
		\end{aligned}
	\end{equation*}
   where $\bm{\mu} = (\iq +\jq +\kq )/\sqrt{3}$ referring to the grey-value axis, and $\bm{\nu}$ is a unit and pure quaternio number and $\bm{\nu}$ is orthogonal to $\bm{\mu}$.
   With the matrix $ C $ defined in \cite{jnw19a}, we can explicitly represent the S-component and V-component of HSV with $ u_r,u_g,u_b $. We have described the specifics on page 3 of the main body. On the other hand, we consider combining the advantages of RGB and HSV spaces to improve image processing. The quaternion can be used as a unified framework to integrate information from different color spaces to enhance image restoration.

\subsection{The dual form of SVTV regularization}

The dual form of the SVTV regularization is provided in \cite{jnw19a}, along with a summary of the SVTV regularization's convexity, lower semi-continuity, approximation, and compactness characteristics. These characteristics are informative in demonstrating the existence of the model solution put out in this work.
In \cite{jnw19a}, an image recovery model is constructed using the SVTV regularization described above, and  the Euler--Lagrange equation of this model is given in terms of Gradient Descent Flow:
\begin{equation*}
	\begin{aligned}
		\nabla\cdot\Bigg(\frac{\nabla\left(2u_{c_1}-u_{c_2}-u_{c_3} \right) }{\sqrt{(|\partial_x\uq(x,y)|_s^2)+(|\partial_y\uq(x,y)|_s^2)}}\\
		+\alpha\frac{\nabla\left( u_r+u_g+u_b\right) }{\sqrt{(|\partial_x\uq(x,y)|_v^2)+(|\partial_y\uq(x,y)|_v^2)}} \Bigg)\\ -3\lambda\left( \hat{K}\star u_{c_1}- z_{c_1}\right)\star \hat{K}^{*}=0,
	\end{aligned}
\end{equation*}
where $c_1,c_2,c_3$ is a sort of cyclic arrangement of $r, g, b$ (There are three arrangements in total: $r, g, b; g, b, r; b, r, g$). It can be seen from these equations that each channel is not calculated independently. Rather, the three channels are coupled together and the information from all three channels is used simultaneously, which ensures the effectiveness of the cross-channel deblurring model proposed later.

We refer to \cite{jnw19a} for the details of transform from quaternion functional to real one.  

\section{Working principle of quaternion blur operator}

\begin{figure}[htbp]
	\centering
	\includegraphics[width=0.32\textwidth]{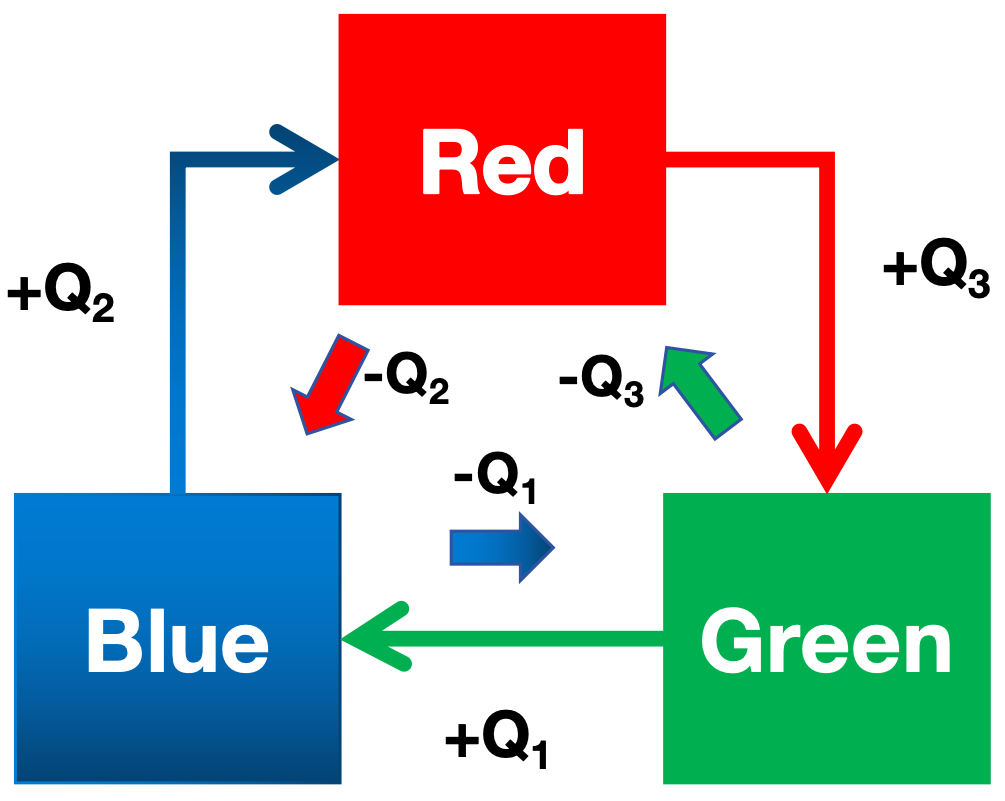}
	\caption{Working principle of $\Qq$.  
	}
	\label{fig:Qblura}
\end{figure}

The quaternion blur operator $\Qq$ in Definition 2 of the main body depicts the rendering between color channels. Its  working principle is shown in Figure \ref{fig:Qblura}. 

\end{document}